\documentclass[10pt,journal,cspaper,compsoc]{IEEEtran}
\usepackage{multicol}
\usepackage[nocompress]{cite}
\usepackage{times}
\usepackage{booktabs}
\usepackage{multirow}
\usepackage{color,soul}
\usepackage{url}
\usepackage[normalem]{ulem}
\usepackage{enumerate}
\usepackage{latexsym}
\usepackage[cmex10]{amsmath}
\usepackage{amssymb}
\usepackage{stmaryrd}
\usepackage{psfrag}
\usepackage{pstricks}
\usepackage{pst-node}
\usepackage{pst-tree}
\usepackage{array}
\usepackage{mdwmath}
\usepackage{mdwtab}
\usepackage{eqparbox}
\usepackage{amssymb}
\usepackage[pdftex]{graphicx}
\usepackage{latexsym}
\usepackage{esvect}
\usepackage{amsfonts}
\usepackage{amsmath}
\usepackage{bbm}

\usepackage{mathtools}
\usepackage{float}
\usepackage{algorithmic}
\usepackage[boxed]{algorithm}
\usepackage{balance}
\usepackage{comment}
\usepackage{subcaption}
\usepackage{cleveref}
\graphicspath{{./Images/}}
\newcommand\norm[1]{\lVert#1\rVert}

\begin{document}

\title{Tracking People in Highly Dynamic Industrial Environments}

\author{Savvas~Papaioannou,~Andrew~Markham~and~Niki~Trigoni

\IEEEcompsocitemizethanks{\IEEEcompsocthanksitem The authors are with the Department of Computer Science, University of Oxford, Oxford, OX1 3QD, UK.\protect\\
E-mail: firstname.lastname@cs.ox.ac.uk}
\thanks{}}

\markboth{IEEE TRANSACTIONS ON MOBILE COMPUTING, VOL. 16, NO. 8, AUGUST 2017}%
{Papaioannou \MakeLowercase{\textit{et al.}}: Accurate Positioning via Cross-Modality Training}

\IEEEcompsoctitleabstractindextext{%
\begin{abstract}
To date, the majority of positioning systems have been designed to operate within environments that have long-term stable macro-structure with potential small-scale dynamics. These assumptions allow the existing positioning systems to produce and utilize stable maps. However, in highly dynamic industrial settings these assumptions are no longer valid and the task of tracking people is more challenging due to the rapid large-scale changes in structure. In this paper we propose a novel positioning system for tracking people in highly dynamic industrial environments, such as construction sites. The proposed system leverages the existing CCTV camera infrastructure found in many industrial settings along with radio and inertial sensors within each worker's mobile phone to accurately track multiple people. This multi-target multi-sensor tracking framework also allows our system to use cross-modality training in order to deal with the environment dynamics. In particular, we show how our system uses cross-modality training in order to automatically keep track environmental changes (i.e. new walls) by utilizing occlusion maps. In addition, we show how these maps can be used in conjunction with social forces to accurately predict human motion and increase the tracking accuracy. We have conducted extensive real-world experiments in a construction site showing significant accuracy improvement via cross-modality training and the use of social forces. 
\end{abstract}

}

\maketitle

\IEEEdisplaynotcompsoctitleabstractindextext
\IEEEpeerreviewmaketitle

\IEEEraisesectionheading{\section{Introduction}\label{sec:introduction}}

\IEEEPARstart{I}{n} today's large and complex industrial environments such as construction sites the need of advanced planning and scheduling, careful coordination, efficient communication and reliable activity monitoring is essential for productivity and safety purposes. Accurate and cost effective positioning and identification are the two main key requirements in order to meet all the above goals. Although positioning technologies have reached a significant level of maturity over the last years there is still no adequate solution for providing accurate positioning services across large and complex industrial settings. 

More specifically, tracking the workers in a construction site is much more challenging than indoor positioning mainly due to the many moving parts and the fast large-scale changes that occur in these complex environments. For instance, in an indoor environment, the positions of walls and floors remain constant over time, whereas positions of furniture vary little from day to day. Existing indoor positioning systems leverage this environmental stability to provide accurate location services with the use of stable maps. In contrast, the construction site evolves rapidly from day to day, precluding the use of systems which rely on stable, long-term maps for positioning. Currently, there is no system that allows for workers to be tracked reliably and robustly during all phases of construction. As a case in point, consider the challenges in a unified positioning system that works equally as well during deep foundation excavation through to an almost complete multi-storey building. At different points in time, the performance of different techniques alters, with some improving and some degrading. 
\begin{figure}
	\centering
	\includegraphics[width=\columnwidth]{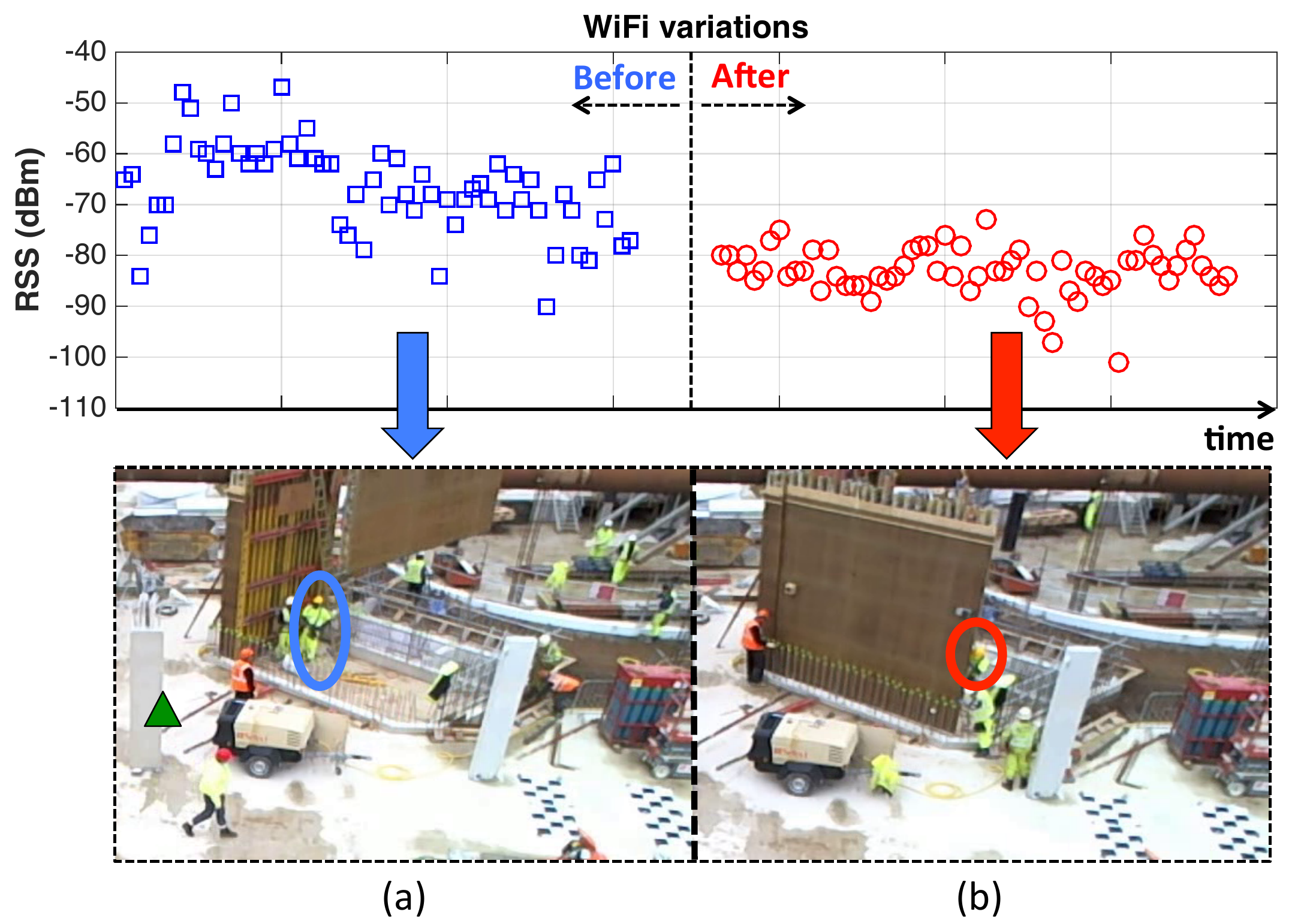}
	\caption{The WiFi signal strength received by the worker in circle is affected by the installation of a new wall (a) Before the installation, there are direct WiFi signals from the access point (shown as triangle) to the worker,  (b) The worker is blocked by the new wall, which affects the propagation properties of the WiFi signals as shown in the graph above. }
	\label{fig:wall}
	\vspace{-3mm}
\end{figure}

\begin{figure}
	\centering
	\begin{subfigure}{0.48\columnwidth}
		\includegraphics[width=\columnwidth]{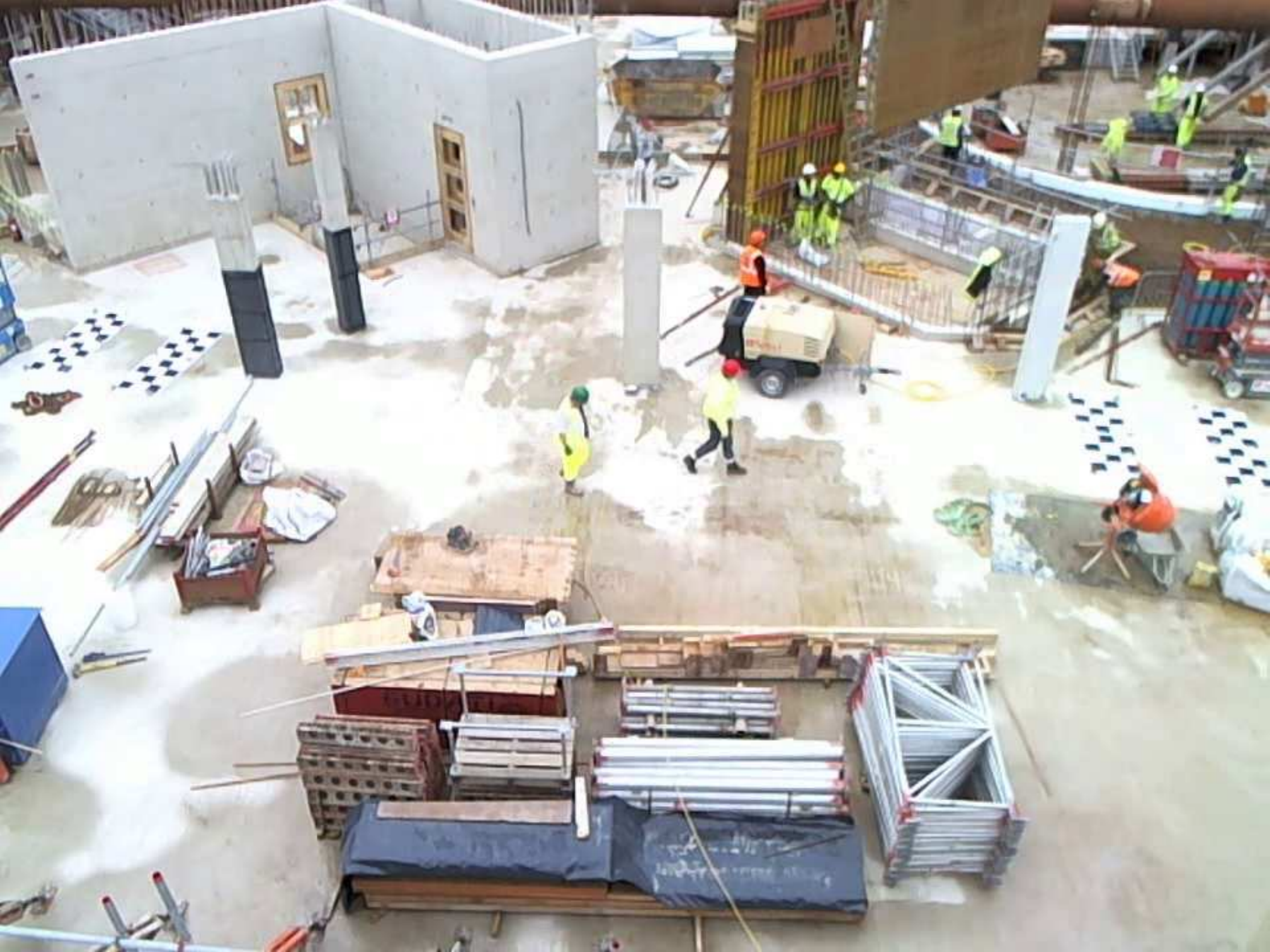}%
	\end{subfigure}\hfill%
	\begin{subfigure}{0.48\columnwidth}
		\includegraphics[width=\columnwidth]{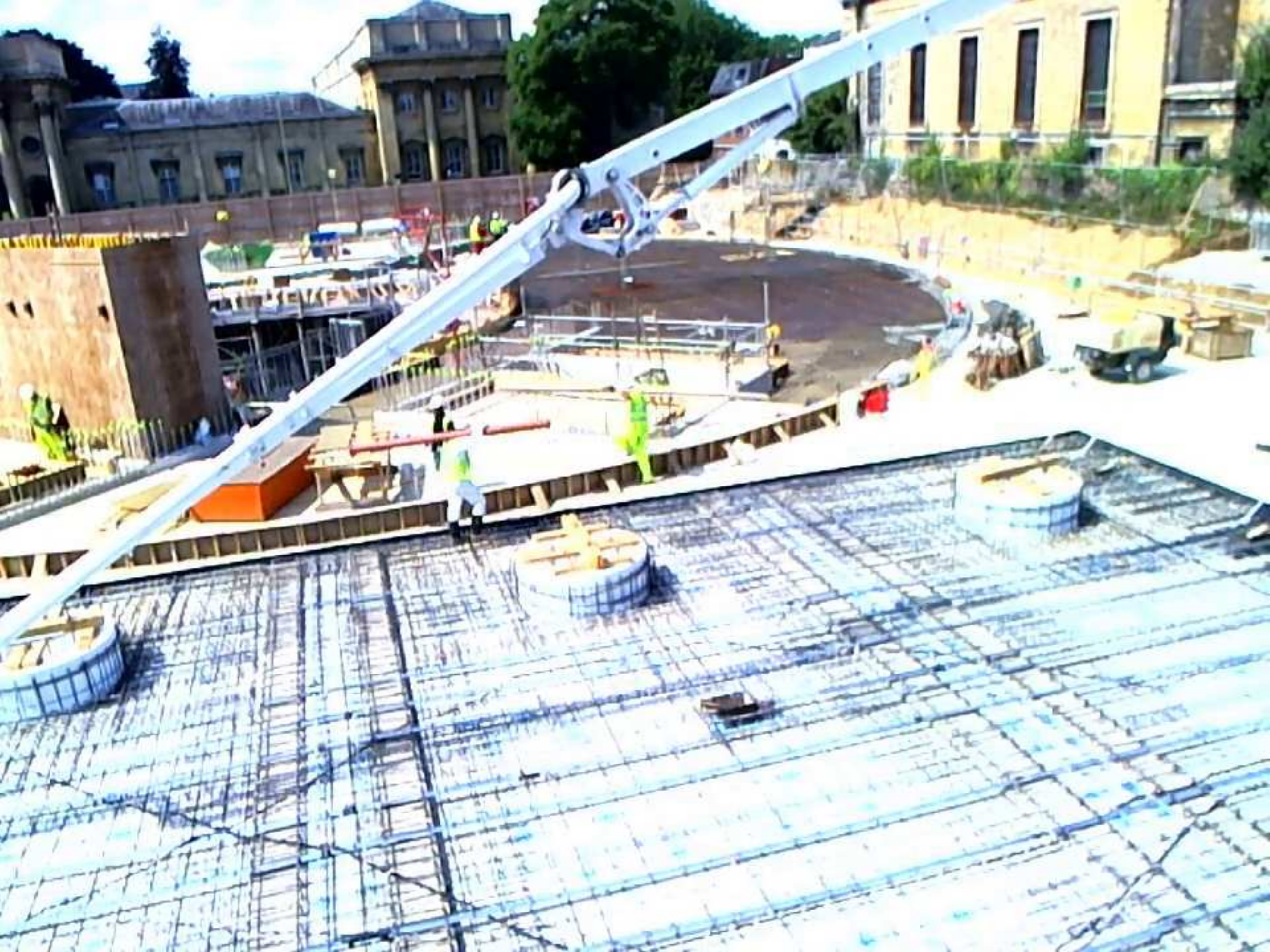}%
	\end{subfigure}\hfill%
	\caption{We have conducted tracking experiments in a construction site setting: (a) the construction site on day 1, (b) the construction site on day 36. The site changes rapidly from day to day, precluding the use of positioning systems which rely on stable, long-term maps.}
	\label{fig:expSetup}
\end{figure}

In this paper, we propose a multi-sensor tracking system which makes use of visual, radio and inertial measurements in order to tackle the problem of accurate localization and identification in construction sites which are characterized by rapid large-scale changes in structure. For example, Fig. (\ref{fig:wall})  shows the effect of a wall being installed in the middle of one of our tracking experiments. The received signal strength of a worker's smartphone from one of the access points dropped considerably after the installation of the wall, in a matter of minutes. The field of view of the camera also changed, not allowing us to directly visually track the people behind the wall. In addition to these short changes, during our experiments we observed much more significant long term changes (Fig. (\ref{fig:expSetup})); within periods of a few weeks, the scene changed dramatically, staircases or entire floors were added, obfuscating the view to the first floors and creating additional layers where people needed to be tracked.  Moreover, the radio and magnetic maps proved unstable with the movement of large structures and the uniforms that people wear for safety make them very hard to distinguish visually, necessitating the use of a multi-sensor tracking framework.


Our aim is to provide a system that can monitor the location of workers to indicate working hazards (e.g. red and green zones), which can be individually tailored. For example, a steel-worker has the training to operate in areas which might not yet be poured with concrete whilst forming the steel rebar. Conversely, a general construction worker should not venture into regions where steel-work has not been completed. This level of safety requires positioning precision beyond the majority of indoor positioning solutions, with desired sub-meter accuracy.

In essence, we are exploiting the fact that different sensing technologies have uncorrelated failure modes to provide a robust, adaptive positioning framework.
To summarize, the major contributions of this work are as follows:
\begin{enumerate}
	\item We are investigating the problem of tracking in highly dynamic industrial settings and we a are proposing a positioning framework explicitly designed for these rapidly changing environments. Our particle-filter based multi-hypothesis tracking framework utilizes three different sensor modalities (i.e. vision, radio and inertial) to allow for accurate tracking in challenging conditions and environments such as construction sites which are characterized by rapid large-scale changes in structure.
	\item A technique for cross-modal sensor parameter learning. The proposed system is able to automatically tune the parameters of its sub-systems (e.g. radio model, visual detector, step-length) by making use of the tracking output and a subset of sensor modalities.
	\item We demonstrate the impact of applying the social force model to improve tracking in dynamic environments. In a construction site the environment changes rapidly with the addition of new walls, corridors, etc. These changes define the walkable area by restricting human motion in certain locations. In this work we show how to take advantage of these environmental changes with social forces to significantly increase the tracking accuracy. 
	\item We have conducted extensive experiments in a real construction site with the help and guidance of our industrial partners.\\
\end{enumerate}

\section{Problem Definition}
\label{sec:problem}
In this paper we tackle the problem of tracking people in environments equipped with one or more stationary calibrated cameras. We assume that people that desire to be tracked carry a mobile device, such as a smartphone or customized worker safety equipment, and move freely in and out of the field of view (FOV). We divide time into short time intervals, and at each time $t$ we receive a number of camera detections of the moving objects denoted as $C_t=\{ c^1_t,c^2_t, ... ,c^j_t, ... \}, ~1 \le j \le |C_t|$. A camera detection $c^j_t$ represents the bounding box of the $j_\text{th}$ object generated by a foreground detector. Note that at time $t$ we could be receiving camera detections not only from people but also from other moving objects (i.e. vehicles); false positive detections are also received due to illumination changes, shadows, etc. In order to reduce the number of false positive detections and concentrate on detecting only people we apply a head detector to the output of a foreground detector. A camera detection $c^j_t$ is projected into the ground plane via a projective transformation which will be denoted as $\hat{c}^j_t$ in this paper.

\noindent At time $t$ we also receive a collection of radio measurements $R_t=\{ r^k_t \}, ~1 \le k \le K$ where $K$ is the total number of people with mobile devices who wish to be tracked and $r^k_t=[ \text{rss}^1, ..., \text{rss}^m  ]^k_t$ is a vector of received signal strength (RSS) measurements of the $k_\text{th}$ device from $m$ access points. Additionally, we assume that each mobile device is equipped with an inertial measurement unit (IMU) containing an accelerometer and a magnetometer. This allows us to generate at time $t$ a collection of inertial measurements denoted as $S_t=\{s^k_t\} $ where $s^k_t=[b^k_t, d^k_t, \theta^k_t]$ is a vector that contains the step indicator, step-length and heading of the $k_\text{th}$ person respectively. Each index $k$ uniquely identifies a person and corresponds to a unique MAC address of the mobile device.

The problem to solve is the following: \emph{Given anonymous camera detections $C_{1:t}$, id-linked radio measurements $R_{1:t}$ and id-linked inertial measurements $S_{1:t}$ estimate the trajectories of all users carrying mobile devices and moving inside the camera FOV}.

\section{System Overview}

\begin{figure}
	\centering
	\includegraphics[width=\columnwidth]{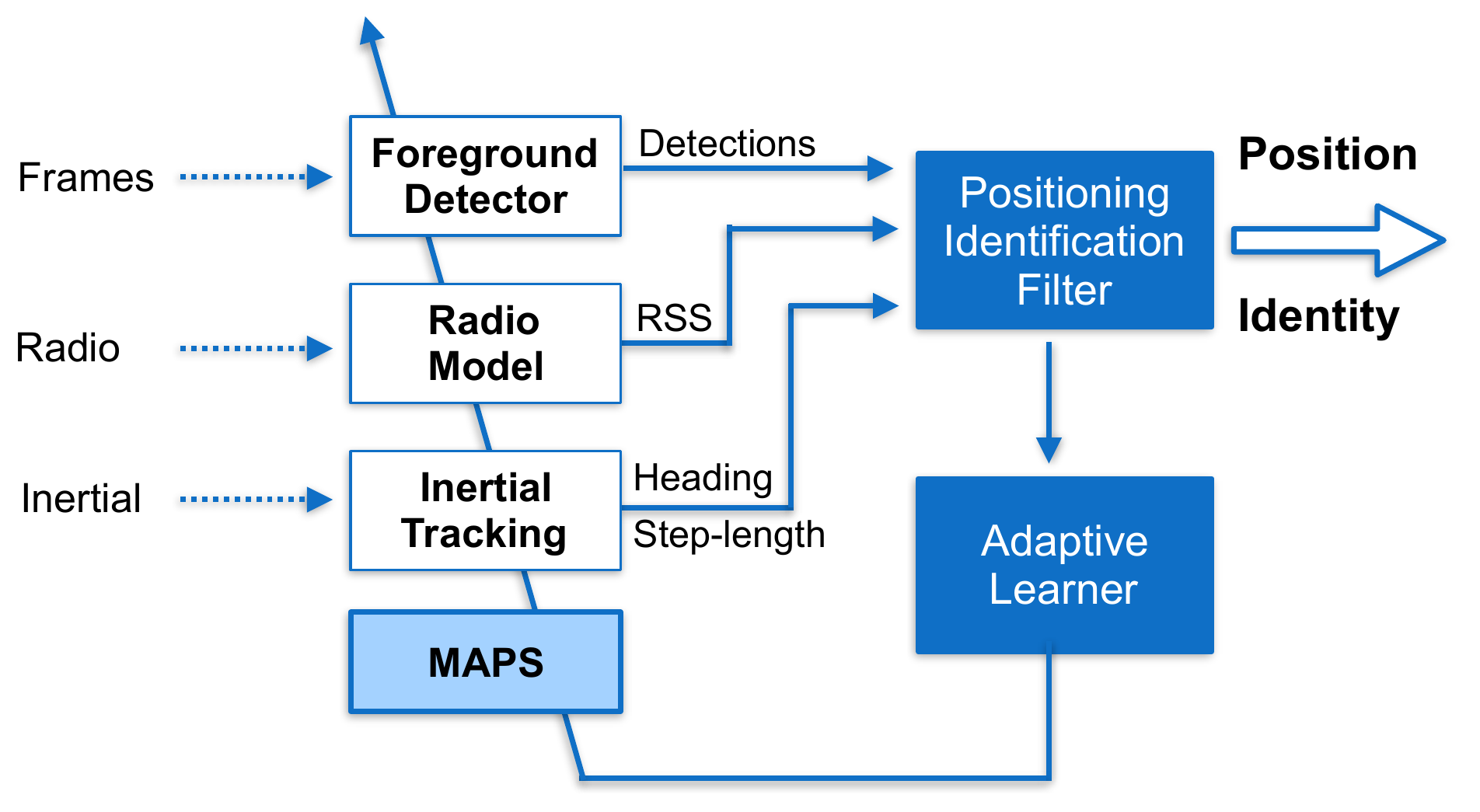}
	\caption{Overview of the proposed system architecture.}
	\label{fig:arch}
\end{figure}

An overview of the proposed system architecture is shown in Fig.~(\ref{fig:arch}). The \emph{Positioning and Identification filter} obtains anonymous camera detections, radio and inertial measurements from multiple people and is responsible for solving three problems. Firstly, it establishes the correspondences of camera detections across frames, that is, it links together anonymous camera detections that correspond to the same person. Secondly, it finds the mapping between anonymous camera detections and id-linked smartphone (radio and step) measurements. Finally, it identifies and estimates the positions of multiple targets. 

\noindent The \emph{Adaptive Learner} uses the output of the filter in combination with the input observations, and performs cross-modality training. Specifically, it configures the foreground detector's internal parameters taking into account available motion measurements. In addition, it tunes the step-length estimation method by leveraging reliable camera measurements.  Finally, it exploits camera measurements to learn the radio model; radio, magnetic and occlusion maps can also be learned which can be used to further improve the system's accuracy. The remaining components of the system are existing modules which pre-process raw sensor data and transform them to camera, step and radio measurements.

\section{Multiple Target Tracking} \label{sec:MTT}
%


In this section we provide a brief overview of previous work on multiple target tracking (MTT). A more detailed description of MTT algorithms can be found in \cite{Blackman1999}.

\subsection{Introduction to Multiple Target Tracking}
Under the general MTT setup a number of indistinguishable targets are assumed to move freely inside the field of view; they can enter and exit the FOV at random times. The system receives sensor data about the position of the targets periodically which are noisy, include false alarm measurements (i.e. background noise or clutter) and occur with some detection probability. Each target follows a sequence of states (e.g. positions) during its lifetime called \emph{track}. 
The main objective of MTT is to collect sensor data containing multiple potential targets of interest and to then find the tracks of all targets and filter out the false alarm measurements. If the sequence of measurements associated with each target is known (i.e. id-linked measurements) then the MTT reduces to a state estimation problem (e.g. distinct Kalman/particle filters can be used to follow each target). However, when the target-to-measurements association is unknown (for example, anonymous measurements from cameras, radars and sonars are used) the data association problem must be solved in addition to state estimation. Essentially, the data association problem seeks to find which measurements correspond to each target.

\subsection{Rao-Blackwellized Particle Filtering}
The main idea of Rao-Blackwellized particle filtering (RBPF) \cite{Doucet2000,doucet2001} is to reduce the number of variables that are sampled by evaluating some parts of the filtering equations analytically. This reduction makes RBPF computationally more efficient than the standard particle filter, especially in high dimensional state-spaces. 

The Rao-Blackwellized Monte Carlo Data Association filter (RBMCDA) \cite{sarkka2004,sarkka2007} is a  sequential Monte Carlo MTT method that uses Rao-Blackwellized particle filtering (RBPF) to estimate the posterior distribution of states and data associations efficiently. More specifically, instead of using a pure particle representation of the joint posterior distribution of states and data associations 
RBMCDA proceeds by decomposing the problem into two parts: a) estimation of the data-association posterior distribution and b) estimation of the posterior distribution of target states. The first part is estimated by particle filtering and the second part is computed analytically using Kalman filtering 
The aforementioned decomposition is possible, since in RBMCDA the dynamic and measurement model of the targets are modeled as linear Gaussian conditioned on the data association thus can be handled efficiently by the Kalman filter. 

\begin{algorithm}
	\begin{algorithmic}[1]
		\STATE \textbf{Input:}  $N$ particles, a measurement vector $y_t$.
		\STATE \textbf{Output:} $p(x_t,\lambda_t|y_{1:t})$: the joint distribution of target states and target-to-measurement associations at time $t$ given measurements up to time $t$.
		\FOR{ each particle $i \in (1..N)$ }
		\STATE For all targets run Kalman filter prediction step.
		\STATE Form the importance distribution as:\\
		For all association events $j$ calculate the unnormalized association probabilities:\\
		$\hat{\pi}^{(i)}_j =  \hat{p}(y_t|\lambda^{(i)}_t = j,y_{1:t-1},\lambda^{(i)}_{1:t-1})  p(\lambda^{(i)}_t = j|\lambda^{(i)}_{1:t-1}) $
		\STATE Normalize the importance distribution.
		\STATE Draw new $\lambda^{(i)}_t$ from the importance distribution.
		\STATE Update target $\lambda^{(i)}_t$ with $y_t$ using Kalman correction step.
		\STATE Update particle weight.
		\ENDFOR
		\STATE Normalize particle weights.
		\STATE Resample.
		\STATE Approximate $p(x_t,\lambda_t|y_{1:t})$ as:\\
		$p(x_t,\lambda_t|y_{1:t}) \approx \sum_{i=1}^{N} w^{(i)}_t \delta(\lambda_t - \lambda^{(i)}_t) \mathcal{N}(x_t|M^{(i)}_t,P^{(i)}_t)$ where $(M^{(i)}_t,P^{(i)}_t)$ are the means and covariances of the target states of the $i_\text{th}$ particle.
	\end{algorithmic}
	\caption{A high-level description of the RBMCDA filter}
	\label{alg:rbmcda}
\end{algorithm}

A high level overview of the RBMCDA algorithm is shown in Alg. (\ref{alg:rbmcda}). The algorithm maintains a set of $N$ particles and each particle corresponds to a possible association of anonymous measurements ($y_t$) to tracks. Each particle maintains for each target its current state $x_t$ (e.g. location) and state uncertainty (i.e. posterior distribution $p(x_t|y_{1:t})$). In the first step (line 4), a Kalman filter is used to predict the next state of a target based on its previous state ($p(x_t|y_{1:t-1})$). Then, the algorithm considers associating each anonymous measurement with each one of the targets in the particle and estimates the probability of each candidate association event (lines 5-6). The association events are modeled with the association indicator  $\lambda_t$ (e.g. $(\lambda_t=0) \implies $  clutter association at time $t$, $(\lambda_t=j) \implies$ target $j$ association at time $t$, etc). The association probability $\hat{\pi}_j$ for target $j$ is computed from the measurement likelihood $\hat{p}(y_t|\lambda_t)$ and the prior probability of data associations $p(\lambda_t|\lambda_{t-1})$. By sampling the resulting importance distribution, the algorithm selects only one of the candidate associations (line~7) and updates the state of the respective target with the anonymous measurement (line~8). This is repeated for each anonymous measurement (e.g. for each camera detection in the camera frame). The particle's weight is then updated taking into account its previous weight and the probabilities of selected associations (line~9). Once all particles have been updated and their weights normalized (line~11), they are re-sampled based on their normalized weights (line~12). At the end of each iteration, the positions of the targets are estimated as a weighted average (i.e. mixture of Gaussians) across all particles (line~13).  
Note that the algorithm above allows us to enforce data association constraints. For instance, we can express that each track is updated by at most one visual measurement, by suitably modeling association priors in line 5.
The existing RBMCDA algorithm is designed to work with anonymous observations. In the next section we point out how we extend it in order to exploit radio and inertial observations that are inherently linked to unique device IDs (i.e. MAC addresses).

\section{Proposed Approach}\label{sec:algo}
We are now in a position to describe how we extend the RBMCDA framework to address the identification and tracking problem in a construction site setting. The key difference here is that we introduce id-linked observations in addition to the anonymous camera observations 
This impacts a number of steps in the algorithm above as explained in this section. 

\begin{figure}
	\centering
	\includegraphics[width=\columnwidth]{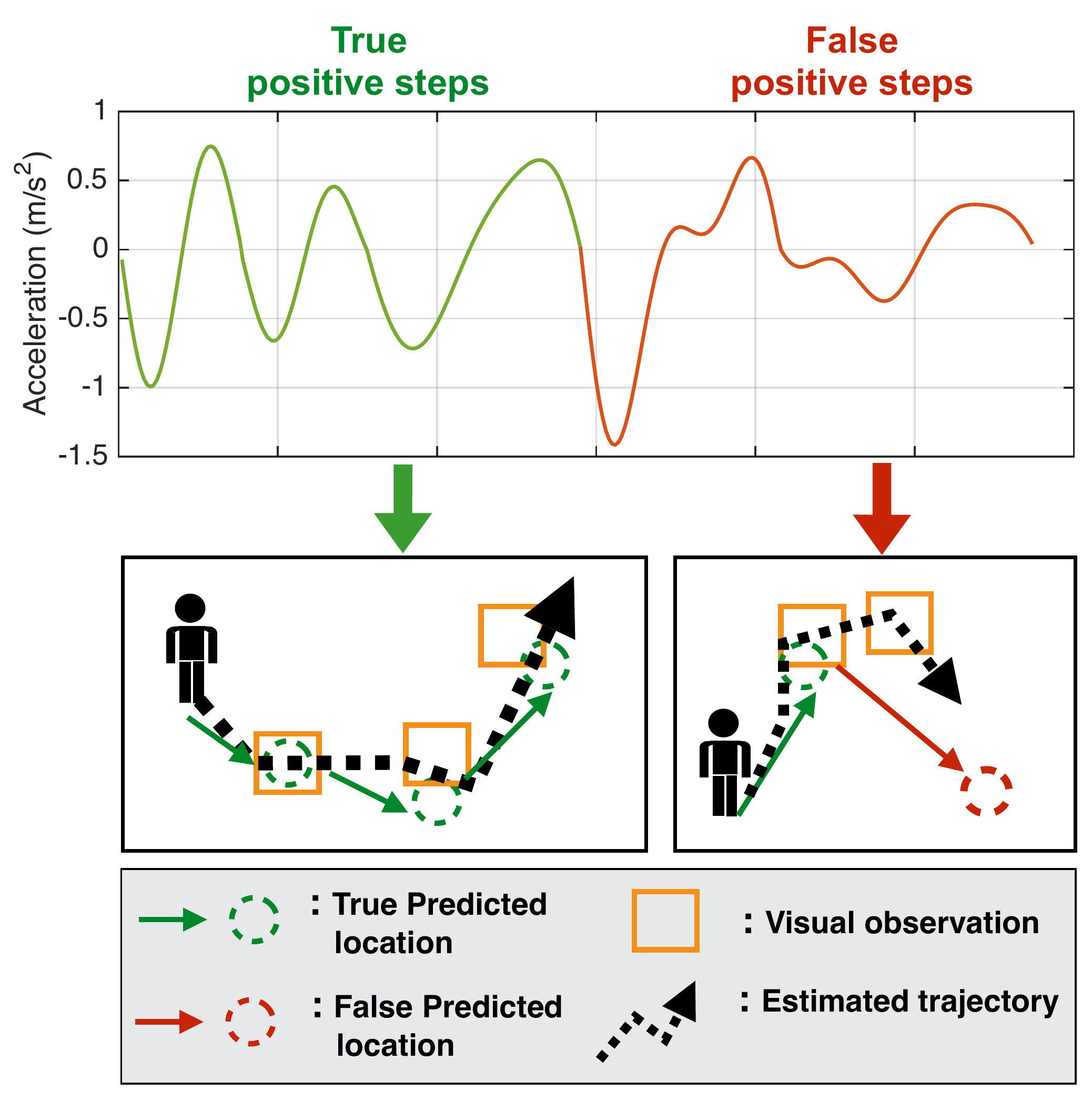}
	\caption{Fusing camera and inertial measurements. The dotted circles show the predicted location using inertial measurements (i.e. a step classifier, shown in the top picture, indicates if a step has been taken or not.) Square boxes indicate a camera detection (i.e. the location of a person). When a step is classified correctly the predicted location is collocated with the camera detection (picture on the left). The tracking accuracy can be decreased significantly when the step detector misclassifies a step. However, in the proposed system, the fusion with camera measurements allows to navigate towards the right path in cases where we have unambiguous trajectories (picture on the right).}
	\label{fig:cameraInertialSteps}
\end{figure}

\subsection{State Prediction and Update}
As in the original algorithm, each particle uses a set of Kalman filters to track targets; however, in our case, we are not interested in tracking all targets within FOV; we only track people equipped with mobile devices and we continue to do so when they temporarily come out of the FOV. We extend the framework in~\cite{sarkka2004,sarkka2007}, in order to use id-linked observations in the prediction and correction steps of the Kalman filter. In particular, we use inertial sensor measurements to predict the next state of a person (instead of only relying on the previous state as in line 4). Furthermore, we use WiFi/BTLE and camera measurements to correct the person's state (instead of only anonymous camera measurements as in line 8). More specifically, the target's dynamics in our system are modeled by the following linear equation:
\begin{equation}\label{eq:xt1}
x_t  =
x_{t-1} + B_t  \begin{bmatrix}
d_{\Delta t} ~ \text{cos}(\theta_{\Delta t})\\
d_{\Delta t} ~ \text{sin}(\theta_{\Delta t})
\end{bmatrix} + w_t
\end{equation}

\noindent where $t$ denotes the time index, $x_t=[x,y]^\text{T}$ is the system state i.e. a $2$-D vector of the target's position on the ground plane and the pair ($d_{\Delta t}$,$\theta_{\Delta t}$) represents the target's step-length and heading respectively calculated within the tracker's cycle time ($\Delta t$). Finally, $B_t$ is a control input indicating whether a step has been taken or not  and $w_t$ is the process noise which is assumed to be normally distributed with mean zero and covariance matrix $\Lambda$ (i.e. $w_t \sim \mathcal{N}(0,\Lambda)$). In order to calculate the step-length of a person we use an empirical model that takes into account the step frequency obtained from the accelerometer data (see Section.~\ref{sec:learning})). In addition, the control input $B_t$ is the output of a HMM-based step classifier which takes as input the accelerometer data from the user's device and returns a step indicator that shows whether a step has been taken or not. A low-pass Butterworth filter (8th order) is being used to smooth out the accelerometer data prior to step classification step. We should also note here that the aforementioned step classifier has classification error of 8.4\% for our dataset. As we already mentioned in Section \ref{sec:problem}, our objective is to track all people that carry mobile devices. Thus, once we associate a camera measurement to a person ID (i.e. device ID), Eqn. (\ref{eq:xt1}) is used as the predictive distribution of a Kalman filter to model the motion of the identified person using his/her inertial measurements. 

Compared with existing techniques (i.e \cite{Papaioannou2014}) that use heuristics to model the human motion, we will show in the evaluation section that the use of inertial measurements in our approach results in more accurate tracking. In addition we have observed that in a construction site workers do not walk regularly, instead they often make big, small and irregular steps depending on the task performed. This makes motion prediction even more challenging since it makes it harder for the step detector/classifier to detect some of the steps correctly. It is worth noting here that the proposed system can correct these step misclassification errors in many situations with the help of visual observations. For instance, when our step classifier predicts wrong for a specific target that a step has been taken, our system can still correct the final estimated position using the location of the camera measurement. Under the assumption of unambiguous tracks the proposed technique can handle similar situations very efficiently. Figure  (\ref{fig:cameraInertialSteps}) illustrates the scenario discussed above.


Unlike the original RBMCDA filter that only uses anonymous observations to update the target's state (line 8), in our system a measurement $y_t$ at time $t$ is a vector containing an anonymous location measurement (2D image coordinates transformed to the world plane via a projective transformation \cite{Hartley2004}) from the camera system and multiple id-linked radio signal strength measurements from people's mobile devices. More formally the measurement vector is defined as $ y_t = [ \hat{c}_t, ~ rss^1_t, ~..., ~ rss^m_t ]^{\text{T}} $ where $\hat{c}_t$ is a camera observation which contains the 2D target coordinates on the ground plane and $rss^1_t, ~..., rss^m_t$ denote the received radio signal measurements from $m$ access-points of a particular mobile device. Thus, the state vector $x_t$ of a target is related to the system measurements $y_t$ according to the following model:

\begin{equation} \label{eq:yt1}
y_t  =
f(x_t) + v_t = \left[
\begin{array}{c}
x_t \\
\text{RSS}_1\left(x_t)\right)  \\
\text{RSS}_2\left(x_t)\right)   \\
\vdots \\
\text{RSS}_m\left(x_t)\right) 
\end{array} 
\right] + v_t
\end{equation}

\noindent where $f$ is a non-linear function that translates the true system state vector to the measurement domain and $v_t$ is the measurement noise which follows a normal distribution with zero mean and covariance matrix $R$ ($v_t \sim \mathcal{N}(0,R)$).  The function $\text{RSS}_i$ is given by:

\begin{equation} \label{eq:rss}
\text{RSS}_i(x_t) = P_i - 10 n_i log_{10} \norm{A_i - x_t}_2 ~,~ i \in [1 .. m]
\end{equation}

\noindent where $m$ is the total number of WiFi/BTLE access points and $\text{RSS}_i(x_t)$ is the expected signal strength at location $x_t$ with respect to transmitter $A_i$. $P_i$ is the received power at the reference distance of 1 meter and $n_i$ is the path loss exponent.
In order to meet the requirements of the RBMCDA filter, i.e. calculate analytically the posterior distribution of the target states with a Kalman filter, Eqn. (\ref{eq:yt1}) must be linear Gaussian. The non-linearity of the measurement model in our case is handled via the unscented transformation \cite{Julier2004}. Thus, the state estimation can be computed analytically using the unscented Kalman filter (UKF) and each particle contains a bank of UKFs; one filter for each target. 
\begin{algorithm}
	\begin{algorithmic}[1]
		\STATE \textbf{Input:} $N$ particles, camera ($C_t$), radio ($R_t$) and inertial ($S_t$) measurements.
		\STATE \textbf{Output:} $p(x_t,\lambda_t|y_{1:t})$.
		\STATE Apply Eqn. (\ref{eq:comb}) to $C_t$ and $R_t$ to create $y_t$.
		\FOR{ each measurement $m \in (1..|y_t|)$}
		\FOR{ each particle $i \in (1..N)$ }
		\STATE For all targets in $i$ run prediction step (Eqn. (\ref{eq:xt1})).
		\STATE Form the importance distribution and draw new association event ($\lambda^{(i)}_t$).
		\STATE Update target $\lambda^{(i)}_t$ with $m$ using UKF correction step. Update particle weight.
		\ENDFOR
		\ENDFOR
		\STATE Normalize particle weights.
		\STATE Resample.
		\STATE Approximate $p(x_t,\lambda_t|y_{1:t})$ as in Algorithm $\ref{alg:rbmcda}$
	\end{algorithmic}
	\caption{A high-level work-flow of the proposed system.}
	\label{alg:rbmcdaMod}
\end{algorithm}

\subsection{Tracking and Identification}
In this section, we show how we modified the association steps in lines 5-7 to leverage id-linked measurements. 

Suppose for instance that at time $t$ we receive camera detections $C_t=\{ c^j_t \}, ~1 \le j \le |C_t|$ and radio measurements  $R_t=\{ r^k_t\}, ~1 \le k \le K$ where $K$ is the number of people with a mobile device. Each one of the $|C_t|$ anonymous camera detections could be one of the following three types: (a) a person with a device, (b) a person without a device or (c) clutter (e.g  false camera detection caused by illumination changes). Our objective is to associate the type (a) camera detections with the correct radio measurements.
In order to do that we follow the following procedure. We enumerate all possible combinations $\Omega=|C_t| \times K$ between the camera detections and the id-linked measurements and we create new measurements $y^i_t , i \in [1..\Omega]$ with the following structure:
\begin{equation}\label{eq:comb}
y^i_t = \{ \hat{c}^m_t , r^j_t  \} , ~ m \in [1..|C_t|],~ j \in [1..K]
\end{equation}

\noindent where $\hat{c}^m_t$ is the camera measurement $c^m_t$ projected into the ground plane. Now, a measurement  $y^i_t$ which contains a correct association will have the following property $\text{RSS}(\hat{c}^m_t) \approx r^j_t$ for the correct $(m,j)$ pair, where $\text{RSS}()$ is the function in Eqn.~(\ref{eq:rss}). In other words, if a person is detected by the camera, then his/her radio measurements (i.e. received signal strength) at that location should match the predicted radio measurements at the same location. Camera detections of type (b) and (c) would normally not exhibit the same property.
From our experiments in a real construction site, we have observed that the radio measurements are reasonably stable but only for short periods of time depending on the environmental dynamics. As we discuss in Section (\ref{sec:learning}) by periodically re-learning the radio model, we make our system adaptive to the changing environment and thus we can use the procedure above to track and identify the people in the scene. 

Moreover, the proposed algorithm can handle the creation and termination of tracks. For instance when a new person (i.e. a new mobile device) is entering the FOV, we initiate a new track by initializing the system state with the camera location that best matches the received radio measurements. Additionally, we allow a target to die when for a fixed period of time no camera observation has been used to update its state. The above procedure runs continuously thus new tracks are created and others are terminated dynamically as people are entering and leaving the FOV. 

As we have already mentioned the association probability is computed as the product of the measurement likelihood and association prior. The measurement likelihood of associating $y^i_t$ with target $j$, $\hat{p}(y^i_t|\lambda_t=j)$ is computed as $\hat{p}(y^i_t|\lambda_t=j)=\mathcal{N}(y^i_t;\hat{y}_{t},V_t)$ where $\hat{y}_{t}$ is the expected measurement of target $j$ at the predicted state and $V_t$ is the innovation covariance obtained from the UKF.

Given $m$ simultaneous measurements within a scan the predictive distribution of data associations can be defined as an $m_\text{th}$ order Markov-chain $p(\lambda^m_t|\lambda^{m-1}_t, . . . , \lambda^1_{t})$ which allows us to enforce certain association restrictions. In our system this predictive distribution is defined (i.e. assigns zero probability to unwanted events) so that the following conditions are met:

\begin{enumerate}
	\item A track can be updated with at most one measurement.
	\item A measurement  can only be used to update at most one track.
	\item An already established track (with a specific sensor ID) can only be updated with a measurement of the same sensor ID.
	\item Once a camera detection is assigned to a track all other measurements which include the latter camera detection are classified as clutter. 
	\item A new target is not born if there is an existing target with the same sensor ID as the newborn target. This means that each particle maintains only targets with unique sensor IDs.
\end{enumerate}
\noindent Some of the above restrictions can be relaxed depending on the application scenario. For instance, when two people are close to each other they can be detected as one object. In this case the 4th restriction can be relaxed in order to allow two tracks (i.e. two people with different sensor IDs) to be updated with the same camera detection.

To summarize, a particle represents states only for people carrying mobile devices - not for all people in the field of view. Inertial data of each person's device are used to predict their next state. Anonymous camera data are associated with a person's track only if they \emph{agree} with both their inertial and radio data. At first a foreground detector is used to detect the moving people in the scene and then the 2D image coordinates of the detected people are projected into the world plane (i.e. ground plane) via a projective transformation (i.e. homography). Given a set of points $p_i$ in the projective plane ${\rm I\!P^2}$ and a corresponding set of points $\hat{p}_i$ likewise in ${\rm I\!P^2}$ we would like to compute the projective transformation that takes each $p_i$ to $\hat{p}_i$. In our case we consider a set of point correspondences $p_i$ $\leftrightarrow$ $\hat{p}_i$ between the image plane and the world ground plane and we need to compute the projective transformation $H_{3 \times 3}$ such that $Hp_i = \hat{p}_i$ for each $i$. The matrix $H$ can be computed using the Direct Linear Transformation (DLT) algorithm \cite{Hartley2004} which requires at least 4 point correspondences. Additional points can improve the estimation by minimizing a suitable cost function such as the geometric distance between where the homography maps a point and where the point's correspondence was originally found, i.e. we would like to find the matrix $H$ which minimizes $\sum_i d(\hat{p}_i, Hp_i)^2$ where $d(.,.)$ is the Euclidean distance between the two points.

Once we calculate $H$ we can use it to project the targets from the image plane into the ground plane and obtain their location on the ground plane. We can then use inertial and radio data using Eqns. (\ref{eq:xt1}) and (\ref{eq:yt1}) as explained earlier in this section.

Finally, we should note here that when at some time-step a particular target does not receive radio measurements then if the target is a new target the identification and creation of a new track is postponed until radio measurements are available. Otherwise, if the target is an existing target, tracking proceeds by only considering the motion model of the target (Eqn. (\ref{eq:xt1})). A high-level work-flow of the proposed technique is shown in Alg.~\ref{alg:rbmcdaMod}).

\section{Cross-modality Learning}\label{sec:learning}
In this section we will show how our framework is capable of cross-modality learning, i.e. how a subset of sensor modalities is used by the \emph{Adaptive Learner} (Fig. (\ref{fig:arch})) to train the internal parameters of the system.

\subsection{Track Quality Estimation}
As we have briefly mentioned in the introduction the output (i.e. track) of our \emph{Positioning and Identification filter} can be used to learn the parameters of various internal components of our system. Once we have identified a track (i.e. we have linked a visual trajectory with radio and inertial measurements), we can use it to learn, for example, the radio propagation model since this track contains all the necessary information (i.e. location-RSS data points) for this purpose. In a similar manner we can learn radio and magnetic maps, train the foreground detector and improve the step-length estimation. All the these will be discussed in more detail later in this section. However, in order to achieve all of the above objectives, we first need to assess the quality of output tracks to make sure that they qualify for the training process. Thus, the goal of the \emph{Track Quality Estimation} phase, is to find candidate tracks which can be used for cross-modality training.

\begin{figure}
	\centering
	\includegraphics[width=\columnwidth]{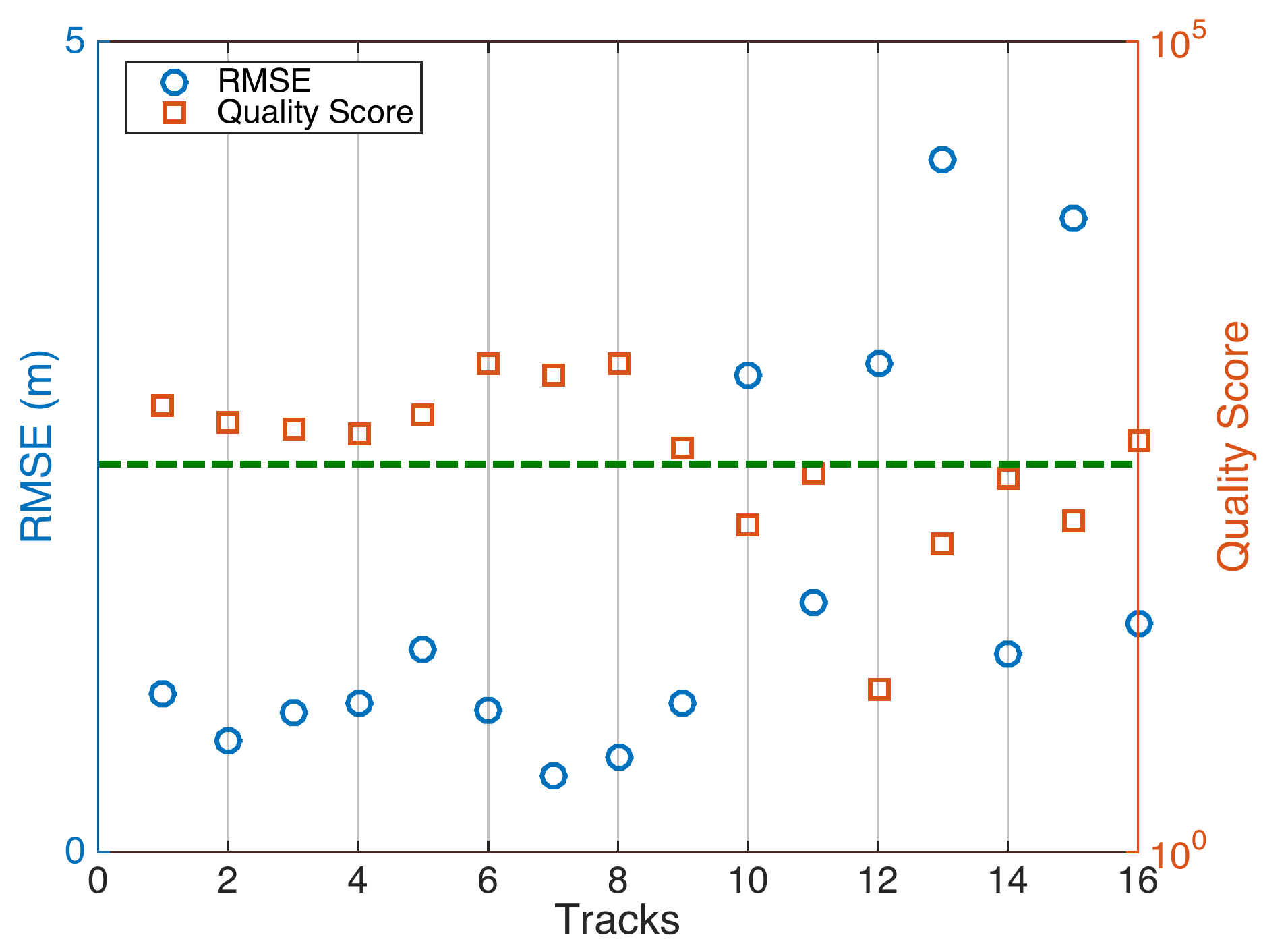}
	\caption{Track quality estimation: The figure shows the quality score of 16 tracks along with their RMSE. Tracks with quality score above the horizontal dotted line are considered qualifying and can be used for cross-modality training}
	\label{fig:quality}
\end{figure}

Let us assume that at time-step (or scan) $t$ we receive $m$ measurements $\{y^1_t,y^2_t,...,y^m_t  \}$. In addition $y^0_t$ is defined for each time-step to be a dummy variable indicating the possibility of a missed detection. Then the incremental quality score of a track $j$ during this time-step is defined as:
\[
\Delta L^j_t = 
\begin{dcases}
\text{log}\left( \frac{\hat{p}(y^i_t|\lambda_t = j) p_d}{\hat{p}(y^i_t|\lambda_t = 0)} \right)  &,~ \text{if}~ \exists ~ i \in [1..m] ~\text{s.t} ~\lambda_t=j\\
\text{log}\left(  1-p_d \right)  &,~  \text{otherwise} 
\end{dcases}
\]

\noindent where the quantity $\hat{p}(y^i_t|\lambda_t = j)$ is the likelihood of the measurement assigned to track $j$. The term $\hat{p}(y^i_t|\lambda_t = 0)=p(clutter)$ is the likelihood of the measurement originating from clutter which has a uniform probability density over the measurement space of volume $V$ (i.e. $p(clutter)=V^{-1}$) and finally $p_d$ is the probability of detection. Then, the cumulative quality score of track $j$ is given by:
\begin{equation}\label{eq:trackScore}
Q_j=\sum\limits_{t=1}^{T} \Delta L^j_t
\end{equation}
where $T$ is the total length of the track. As we can see the quality score $Q$ of a track penalizes the non-assignments due to missing detections while favoring the correct measurement-to-track associations. Fig.~(\ref{fig:quality}) shows that the quality score is negatively correlated with the root mean square error. Finally, in order to mark a track as a high confidence track that \emph{qualifies} for cross-modal training its quality score is tested against a pre-determined threshold $Q_{\text{Th}}$. If $Q_j \ge Q_{\text{Th}}$ then the track is qualified (i.e. \textit{high quality track}) and it can be used for cross-modality training, otherwise the track is rejected (Fig. (\ref{fig:quality})).

\subsection{Foreground Detector Training}
The mixture of Gaussians (MoG)~\cite{Stauffer1999} foreground detection which is used by our system is one of the most popular approaches for detecting moving targets from a static camera.  This approach maintains a statistical representation of the background and can handle  multi-modal background models and slow varying illumination changes. 

In the original algorithm the history of each pixel is modeled by a mixture of $K$ (typically 3-5) Gaussian distributions with parameters ($\beta_k,\mu_k,\sigma_kI $) for the mixture weight, mean and covariance matrix of the $k_\text{th}$ Gaussian component. In order to find the pixels that belong to the background,  the Gaussian distributions are ordered in decreasing order according to the ratio $(\beta_k/\sigma_k)$;  background pixels exhibit higher weights and lower variances than the foreground moving pixels. The background model is obtained as $B^*=\operatorname*{arg\,min}_B \left(  \sum_{k=1}^{B} \beta_k > P_b  \right) $ where $P_b$ is the prior probability of the background. The remaining $K-B^*$ distributions represent the foreground model.

On the arrival of a new frame each pixel is tested against the Gaussian mixture model and if a match is found the pixel is classified as a background or foreground depending on which Gaussian component it was matched with. If no match is found the pixel is classified as a foreground and it is added to the mixture model by evicting the component with the lowest weight. When a pixel is matched, the weight of that $k_{th}$ Gaussian component is updated using an exponential weighting scheme with learning rate $\alpha$ as $\beta_{t+1}=(1-\alpha)\beta_t + \alpha$, and the weights of all other components are changed to $\beta_{t+1}=(1-\alpha) \beta_t$. A similar procedure is used to update the mean and covariance of each component in the mixture. 

 The learning rate ($\alpha$) controls the adaptation rate of the algorithm to changes (i.e. illumination changes, speed of incorporating static targets into the background) and is the most critical parameter of the algorithm.  Fast learning rates will give greater weight to recent changes and make the algorithm more responsive to sudden changes. However, this can cause the MoG model to become quickly dominated by a single component which affects the algorithm's stability. On the other hand slow learning rates will cause a slower adaptation change which often results in pixel misclassification. Over the years many improvements have been suggested by the research community that allow for automatic initialization and better maintenance of the MoG parameters \cite{bouwmans2008}. More recent techniques \cite{Sobral20144,Munir2014} address challenges like sudden illumination variations, shadow detection and removal, automatic parameter selection, better execution time, etc .

In this section we propose a novel method for obtaining the optimum learning rate $\alpha^*$ of the foreground detector using the \emph{high-quality} tracks of our filter. Suppose we are given a  track $X^j_{1:T}=\{x^j_1,x^j_2,...,x^j_T\}$ of length $T$ where $x^j_t, t \in [1..T]$ denotes the state of the track at time $t$. Since, both camera and inertial measurements could have been used to estimate track  $X^j_{1:T}$ then its states $x^j_t, t \in [1..T]$ are of two types: type (a) states that have been estimated using camera and inertial measurements and type (b) states that have been estimated only using inertial measurements. A high-quality track ensures that  $X^j_{1:T}$ contains the right mixture of type (a) and type (b) states and thus does not deviate significantly from the ground truth trajectory. This is possible, since propagating a track by only using inertial measurements is accurate enough for short periods of time. This key property of the inertial measurements allows us to use a high quality track as if it was the ground truth trajectory to train the learning rate of the foreground detector. In other words the type (b) states of a high quality track tells us that the target is moving to specific locations and the foreground detector does not detect any target at those locations. 

The quality score of tracks (Eqn. (\ref{eq:trackScore})) can be used to find the optimum learning rate by solving the following optimization problem: \emph{Given a time window $\mathcal{T}$ find a learning rate $\alpha^*$ so that the cumulative quality score (CQS) $\sum_j Q_j$ of all high quality tracks $j \in \mathcal{T}$ is maximized}.

 \subsection{Optimizing the Step Length Estimation}
Similar to the foreground detector training procedure, \emph{high quality} tracks can also be used to learn the step-length model of each person being tracked. More specifically, the step-length of a user can be obtained from the universal model proposed in \cite{Renaudin2012} as:
 \begin{equation}\label{eq:stepModel1}
 s=h  (a^\prime f_{step} + b^\prime) + c^\prime
 \end{equation}
 where $s$ is the estimated step-length, $h$ denotes the user's height, $f_{step}$ is the step frequency obtained from the device's accelerometer and $(a^\prime, b^\prime, c^\prime)$ are the model parameters. The model above describes a linear relationship between step-length and step frequency weighted by the user's height. 
 \begin{figure}
 	\centering
 	\includegraphics[width=\columnwidth]{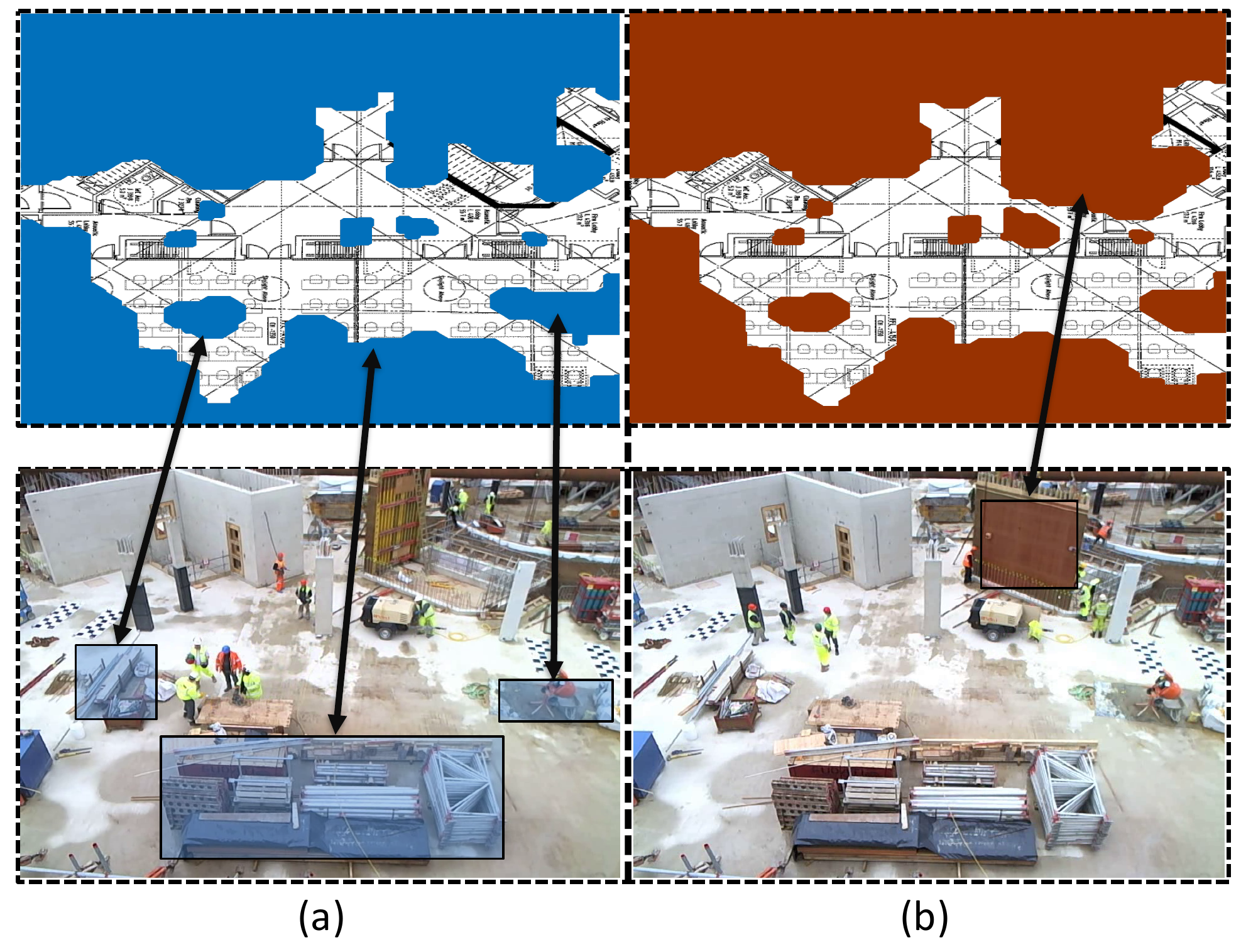}
 	\caption{The figure shows the occlusion maps learned during a period of 10 minutes for each map. (a) Areas that appear to have no human activity are marked as occlusions, (b) As the constuction site evolves new occlusions are created. In this case the installation of a new wall creates a new occlusion. These changes are detected automatically by our system and are used to improve the tracking accuracy via the use of social forces.}
 	\label{fig:occlusionMap}
 \end{figure}
 Since the heights of people that we need to track are not known a priori  every time a new track is initialized that contains a sensor ID which has not been recorded before, the step-length estimator uses Eqn. (\ref{eq:stepModel1}) to provide an initial estimate of the target's step-length. At this point the height value is set to the country's average for men of ages between 25 and 34 years old. The parameters $(a^\prime, b^\prime, c^\prime)$ have been pre-computed with a training set of 8 people of known heights using foot mounted IMUs.
 
 As the tracking process proceeds high quality tracks are obtained periodically for each target. From these tracks the following IMU data are extracted for each step: a) step frequency, b) step start-time and c) step end-time. The start/end times of each step obtained from the IMU data are then matched to camera detections in order to obtain the position of the target during those times which are essentially the step-lengths measured from the camera system. Thus, for each target we obtain a collection of $n$ calibration points $\{ S\text{v}^i, f^i_{step} \}^n_{i=1}$ where $S\text{v}^i$ is the visual step-length of the $i_\text{th}$ step and $f^i_{step}$ its frequency obtained from the IMU. The calibration set of each target is then used to train a personal step-length model of the form $S\text{v}=\varrho_1 f_\text{step} +\varrho_0$ using the least squares fitting. Finally, the step-length estimator can switch to the trained model once the least squares goodness of fit $\left(R^2=1-\frac{\text{residual sum squares}}{\text{total sum squares}}\right)$ exceeds a pre-defined threshold.

\subsection{Radio Model/Maps Learning}
\emph{High quality tracks} are also being used in order to learn the parameters of the radio propagation model which our system uses as explained in Section \ref{sec:algo}. More specifically, from a high quality track $X^j_{1:T}=\{x^j_1,x^j_2,...,x^j_T\}$ of length $T$, the type (a) states are extracted. Let us call a type (a) state as $\tilde{x}^j_t$; this state has been estimated using camera, radio and inertial measurements. Thus a collection of type (a) states $S=\{\tilde{x}^j_t : j \in K, t \in \mathcal{T} \}_n$ of length $n$ where $K$ is the total number of people with smartphones and $\mathcal{T}$ is the running time of our filter, contains $n$ pairs of (location, RSS) measurements. Now, this collection of (location, RSS) points can be used to estimate the parameters of the log-normal radio propagation model \cite{Seidel1992} given by Eqn. ({\ref{eq:rss}}) for each access point using least squares fitting.
At regular intervals we re-estimate the radio model parameters based on the most recent portion of collected data. We should note here that the parameters of the radio model are initialized empirically based on a number of studies for different environments \cite{Seidel1992}.

Additionally, we can follow similar procedure to learn radio, magnetic and occlusion maps. The radio and the magnetic maps can be combined and used for localization in situations where the camera is occluded by an obstacle or they can be used in conjunction with the radio model to improve the system's accuracy. Additionally, the occlusion map, which is derived from the camera detections provides statistics about the environment (i.e. frequent visited areas, inaccessible areas, etc) which our system can use to improve its performance. For instance, suppose that a particular person is not detected by the camera during some time and our filter reverts to IMU tracking; the occlusion map can help us filter out impossible trajectories. 

In order to learn the occlusion map we use the following procedure: We first discretize the world plane creating a 2D grid. During a time-window we then project the camera detections into the world plane and we count the number of hits in each cell creating a 2D histogram. The normalized histogram is then thresholded and the cells that are found to be below a predefined threshold are marked as occlusions/obstacles; this is shown in Fig.~(\ref{fig:occlusionMap}). The set of occlusions found $O=\{o_j\}_{j=1}^{N_o}$ are also used to model repulsive forces exerted from the environment onto people; this will be discussed in Section (\ref{sec:socialforce}).

\section{Integration of Social Forces}\label{sec:socialforce}
In this section we describe how we have modified our system to make use of the Social Force Model (SFM)  \cite{Helbing1995,Luber2010} for accurate motion prediction. More specifically the Social Force Model assumes that the behavior of human motion is affected by the motion of other people and also by obstacles from the environment. Thus the SFM aims to describe and predict the behavior of human motion with the introduction of repulsive forces exerted on people by modeling the interactions between people and the influence of the environment on human motion. As we have already mentioned in the previous section our system is able to automatically learn the occlusion map witch contains the location of obstacles and other environmental constrains. This occlusion map is now integrated into the social force model which help us improve the prediction of human motion.

\begin{figure}
	\centering
	\includegraphics[width=\columnwidth]{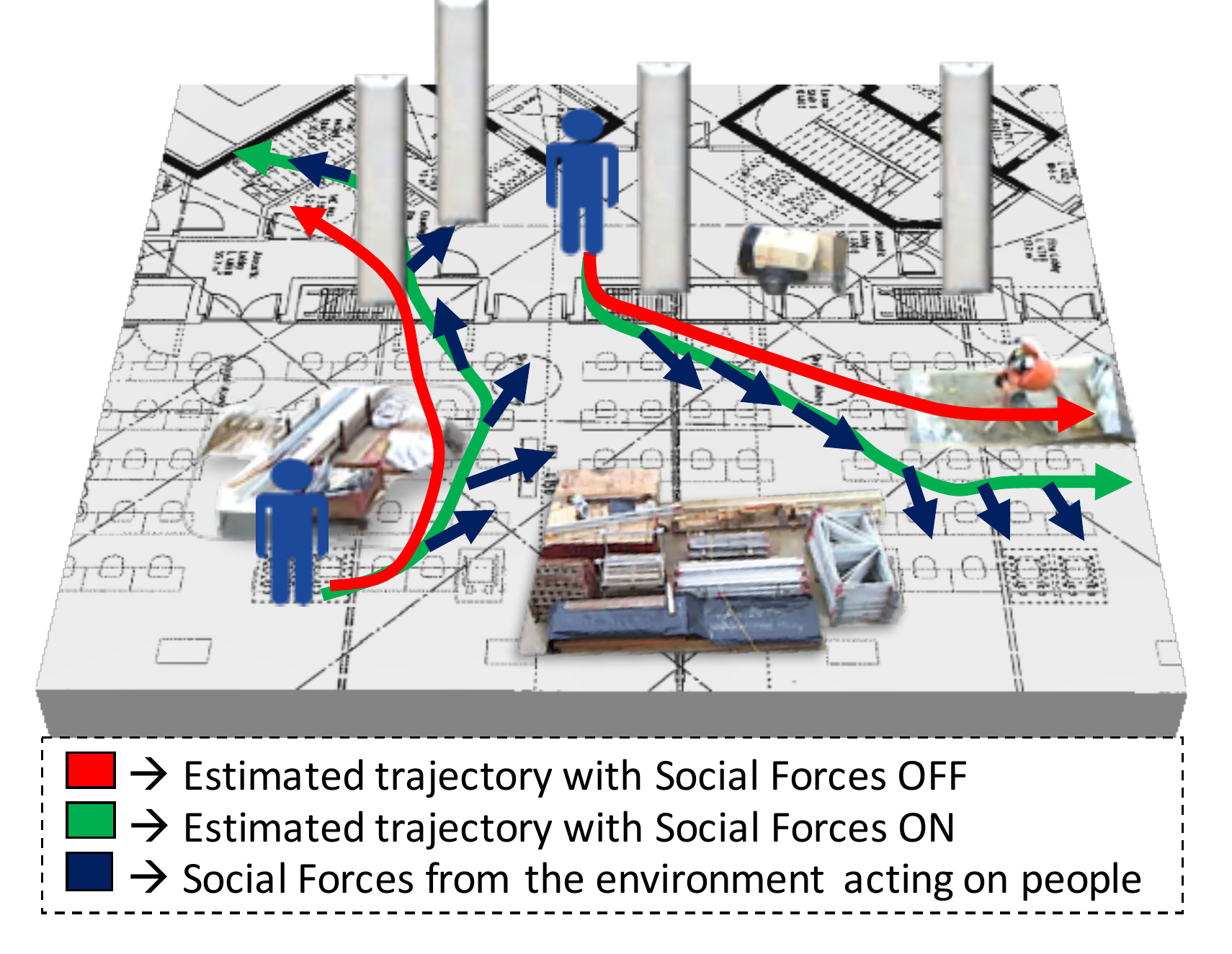}
	\caption{Illustrative example showing the position estimate with and without social forces. The figure shows that
		repulsive physical forces from the environment improve the position estimate by taking into account obstacles and other enviromental constraints.}
	\label{fig:socialForceEx}
\end{figure}

\subsection{The Social Force Model}
More formally in the Social Force Model a person $p_i$ with mass $m_i$ aims to move with a certain desired speed $\hat{\nu}_i$ in a desired direction $\hat{\epsilon}_i$. In our system the desired direction is taken from the IMU measurements (i.e. heading) so that $\hat{\epsilon}_i=\theta^i$ and the desired speed $\hat{\nu}_i$ is calculated as $d_{\Delta t}/\Delta t$ where $d_{\Delta t}$ is the step-length from the IMU and $\Delta t$ the tracker's cycle time.
At each time step the motion of people is described by the superposition of repulsive and physical forces exerted from other people and the environment.


\subsubsection{Repulsive Forces}
As we already mentioned the human motion is affected by environmental constrains (i.e. obstacles) and from the motion of other people. Thus in the presence of other people or obstacles a person might not be able to keep the desired direction and speed. These disturbances are described by repulsive forces which prevent a person from moving along the desired direction. More specifically the repulsive force $F^{R}_i$ is modeled as the sum of social forces $f^{\text{soc}}_{i,k}$ exerted by other people or obstacles according to:
\begin{equation}\label{eq:repulsiveForce}
F^{\text{R}}_i= \sum_{j \in P \backslash\{i\}} f^{\text{soc}}_{i,j} + \sum_{j \in O} f^{\text{soc}}_{i,j}
\end{equation}
where $P=\{p_j\}_{j=1}^{N_p}$ is the set of all people (i.e tracks)  and $O=\{o_j\}_{j=1}^{N_o}$ is the set of all environmental constraints (i.e. obstacles). The above social repulsive forces are described as:
\begin{equation}
f^{\text{soc}}_{i,j} = \alpha_j e^{  \left( \frac{r_{i,j}-d_{i,j}}{b_j} \right)  } n_{i,j} \gamma(\lambda,\phi_{i,j})
\end{equation}
where $j \in P \cup O$ and $a_j$, $b_j$ denotes the magnitude and range of the force respectively. People and obstacles are assumed to be circular objects with certain radii, thus $r_{i,j}$ denotes the sum of radii of entities $i$ and $j$ and $d_{i,j}$ is the Euclidean distance between their centers. The term $n_{i,j}$ describes the direction of the force, (normalized vector ) pointing from entity $j$ to entity $i$. Finally, the social forces are limited to the field of view of humans, therefore the anisotropic factor $\gamma(\lambda,\phi_{i,j})$ is added to the model and is given by:
\begin{equation}
\gamma(\lambda,\phi_{i,j}) = \lambda + (1-\lambda) \frac{1+\text{cos}(\phi_{i,j})}{2}
\end{equation}
where $\lambda$ denotes the strength of the anisotropic factor and $\text{cos}(\phi_{i,j})= -  n_{i,j} \cdotp \hat{\epsilon}_i$ is the cosine of the angle between the desired direction and the direction of the force.

\subsubsection{Physical Forces}
Finally, environmental constraints (i.e. walls, obstructions, etc) define the walkable area by restricting human motion in certain locations. These hard constraints can be modeled as physical forces exerted from the environment onto people and can be defined as follows:
\begin{subequations}
\begin{align}
	F^{\text{phys}}_i =&  \sum_{j \in O} f^{\text{phys}}_{i,j} \label{eq:physForce}\\
	f^{\text{phys}}_{i,j} =& c_j g( r_{i,j} - d_{i,j} ) n_{i,j}
\end{align}
\end{subequations}
where $c_j$ denotes the magnitude of the force and $g(x)$ is defined as $g(x)=x$ if $x \ge 0$ and $0$ otherwise, making $g(x)$ a contact force. We should note here that physical forces can also be applied between people if desired (i.e. so that different people would not occupy the same space). This can be done by adding an additional term in Eqn. (\ref{eq:physForce}) to account for forces between people as we did in Eqn. (\ref{eq:repulsiveForce}).

\subsection{Social Forces for Motion Prediction}
The total force $F^{\text{tot}}_i$ exerted on a particular person $p_i$ is the superposition of all repulsive and physical forces given by:
\begin{equation}
F^{\text{tot}}_i= F^{\text{R}}_i + F^{\text{phys}}_i
\end{equation}
We can now include $F^{\text{tot}}_i$ to our motion model (Eqn. (\ref{eq:xt1})) by making use of Newton's second law given by $F^{\text{tot}}_i=m_i \frac{\text{d}\upsilon_i}{\text{d}t}$ so that Eqn. (\ref{eq:xt1}) becomes:
\begin{equation}\label{eq:forceMotion}
x_t  =
x_{t-1} + B_t  \begin{bmatrix}
d_{\Delta t} ~ cos(\theta_{\Delta t})\\
d_{\Delta t} ~ sin(\theta_{\Delta t})
\end{bmatrix} + \frac{1}{2}\frac{F^{\text{tot}}}{m} \Delta t^2 +  w_t
\end{equation}
As we can see from Eqn. (\ref{eq:forceMotion}) the predicted motion of a person is calculated by taking account the previous position, inertial measurements (i.e. step-length and heading), and the forces exerted to this person by other people and the environment. Equation (\ref{eq:forceMotion}) can now  be used in our tracking framework as the predictive distribution of the Kalman filter. This predictive distribution is given by:
\begin{multline}
p(x_t | x_{t-1},S^\prime_t, P_t, O_t) =\\
 \mathcal{N}( x_t ; \psi(x_{t-1},S^\prime_t, P_t, O_t), J_\psi \Sigma _{t-1}   J^{\text{T}}_\psi  + \Lambda )
\end{multline}
where $S^\prime_t=[d_{\Delta t} cos(\theta_{\Delta t}),~d_{\Delta t} sin(\theta_{\Delta t})]^\text{T}$ is a vector that contains the step-length and heading at time $t$, $P_t$ is the set of all people tracked and $O_t$ is the set of all obstacles from the environment. The function $\psi(x_{t-1},S^\prime_t, P_t, O_t)= x_{t-1} + B_t S^\prime_t + \frac{1}{2}\frac{F^{\text{tot}}}{m} \Delta t^2$ is the mean of the predicted location, $\Sigma _{t-1}$ is the covariance matrix of the estimate, $\Lambda$ is the covariance matrix of process noise and $J_\psi=\frac{\partial \psi(\cdot)}{\partial x}$ is the Jacobian of $\psi(\cdot)$.

With the above motion model, in each time step in addition to the inertial measurements we can now use repulsive and physical forces exerted on targets in order to improve the predicted location estimates (Fig. (\ref{fig:socialForceEx})). We will show in the evaluation section that the integration of social forces in our motion model allows us to make better motion predictions and improve the accuracy of our tracking system.

\section{System Evaluation}\label{sec:eval}

\subsection{Experimental Setup}

In order to evaluate the performance of the proposed approach we have conducted two real world experiments in a construction site (Fig. (\ref{fig:expSetup})). In both experiments we placed two cameras with non-overlapping FOV at approximately 8 meters above the ground facing down. In the first experiment the two cameras were covering an area of approximately  11m $\times$ 9m each and in the second experiment an area of 14m $\times$ 4m each. The duration of each of the experiments was approximately 45 minutes with the cameras recording video at 30fps with a resolution of 960 $\times$ 720 px. We should also mention here that each camera was processed separately (i.e. we do not consider the multi-camera system scenario). The area of the site was outfitted with 12 WiFi and 8 BTLE access points and 5 workers were supplied with smartphone devices. The total number of people in the scene was varying from 3 to 12 as workers were entering and exiting the field of view.
The objective of the experiment was to identify and track the workers who were carrying a smartphone device using camera, radio and inertial measurements. The radio measurements were obtained by their smartphones receiving WiFi and BTLE beacons at 1Hz and 10Hz respectively. The inertial measurements (i.e accelerometer and magnetometer) obtained from their smartphones had a sampling rate of 100Hz.

To obtain the ground truth of people's trajectories we followed the same approach proposed in \cite{Papaioannou2014}. We supplied all people to be tracked with helmets of different colors and their ground truth trajectories were obtained using a mean-shift tracker \cite{Ning2012} to track the colored helmets. We have decided to use the procedure above for obtaining the ground truth trajectories since with GPS we could not get the required accuracy (i.e. GPS achieved a room-level accuracy during our experiments at the construction site) for this specific task.

In our implementation we have used RGB images as input to the MoG foreground detector, however we have not used any color features for people identification and our filter tracks only the position of targets. Any target detector which outputs target coordinates can be used with the proposed technique without any changes to the algorithm. It is also worth mentioning that the proposed system can also be extended to utilize visual features (i.e. color) for target identification, however these features are not always available and therefore cannot be relied on to uniquely identify the workers. 
Finally, Table (\ref{table1}) shows all the empirical values and thresholds that we have used in our implementation. These values have been obtained experimentally unless otherwise stated.

\begin{table}[!t]
\renewcommand{\arraystretch}{1.3}
\caption{Empirical Values and Thresholds used in our Implementation}
\label{table1}
\begin{center}
	\begin{tabular}{| l | p{2.2cm} | p{3.5cm} |}
		\hline
		\textbf{Symbol} & \textbf{Description} & \textbf{Values, [Units]}\\
		\hline \hline
		$\Lambda$ & Process noise covariance & diag\{$0.3^2, 0.3^2$\},  
		
		[$\text{m}^2$, $\text{m}^2$]\\ \hline
		$R$ & Measurement noise covariance &  diag\{$0.2^2, 0.2^2, 3.2^2, \ldots$\},
		
		$[\text{m}^2$, $\text{m}^2$, $\text{dBm}^2$, $\ldots$] \\ \hline
		$\alpha$ & MoG learning rate & 0.0032 (learned) \\ \hline
		$P_b$ & MoG prior probability of background & 0.82 \\ \hline
		$K$ & MoG number of Gaussians & 5 \\ \hline
		$Q_{\text{Th}}$ & Quality Score threshold & 300\\ \hline
		$(a^\prime, b^\prime, c^\prime)$ & Step-length empirical model & (0.1244, 0.066, 0.2000) \\ \hline
		$(a_j, b_j, c_j)$ & Social Forces parameters & (50, 0.5, 250),
		
		[N, m, N/m]\\	
		\hline
	\end{tabular}
\end{center}
\end{table}

\begin{figure}
	\centering
	\begin{subfigure}{0.5\columnwidth}
		\includegraphics[width=\columnwidth]{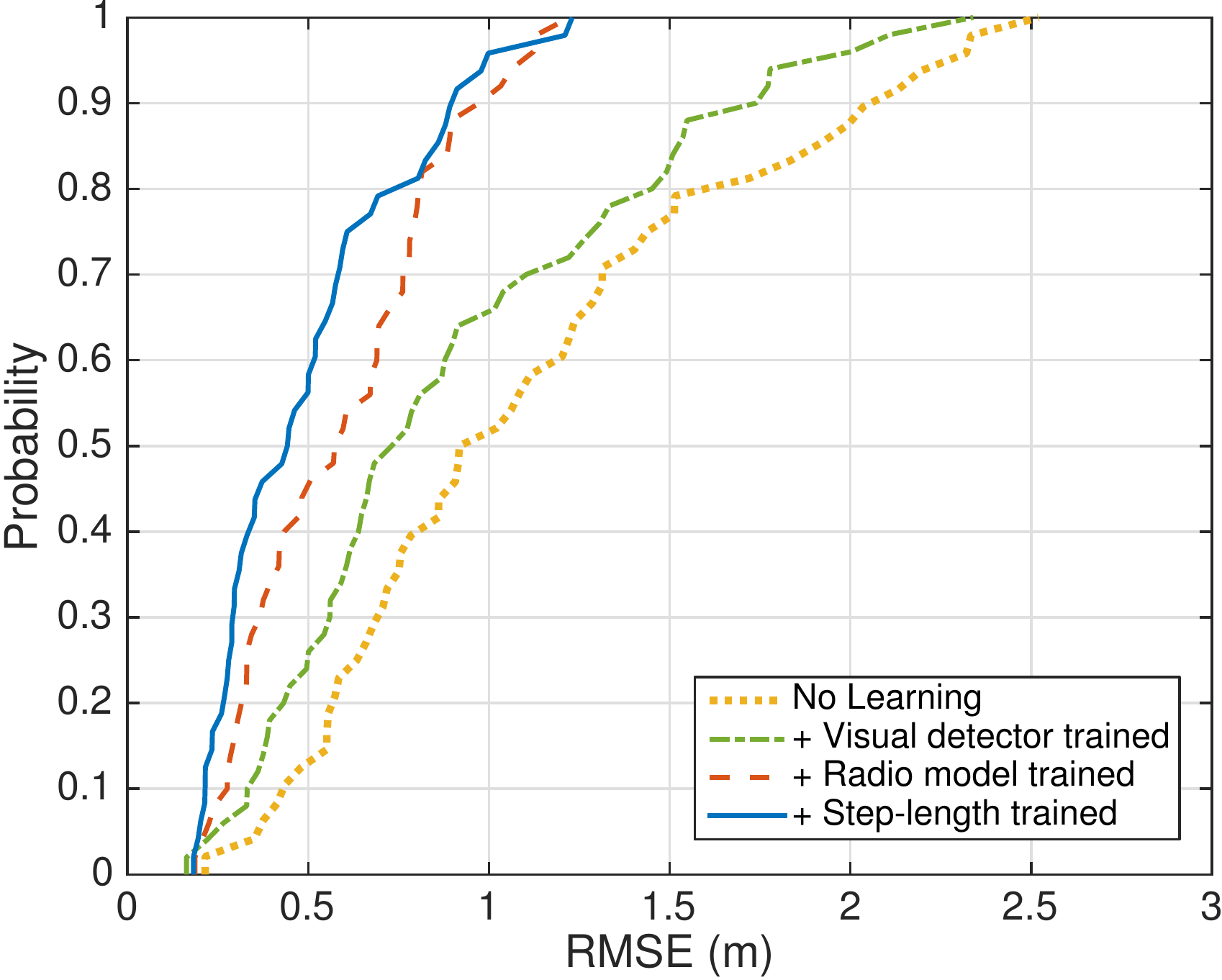}
		\caption{}%
		\label{fig:cdf}%
	\end{subfigure}\hfill%
	\begin{subfigure}{0.5\columnwidth}
		\includegraphics[width=\columnwidth]{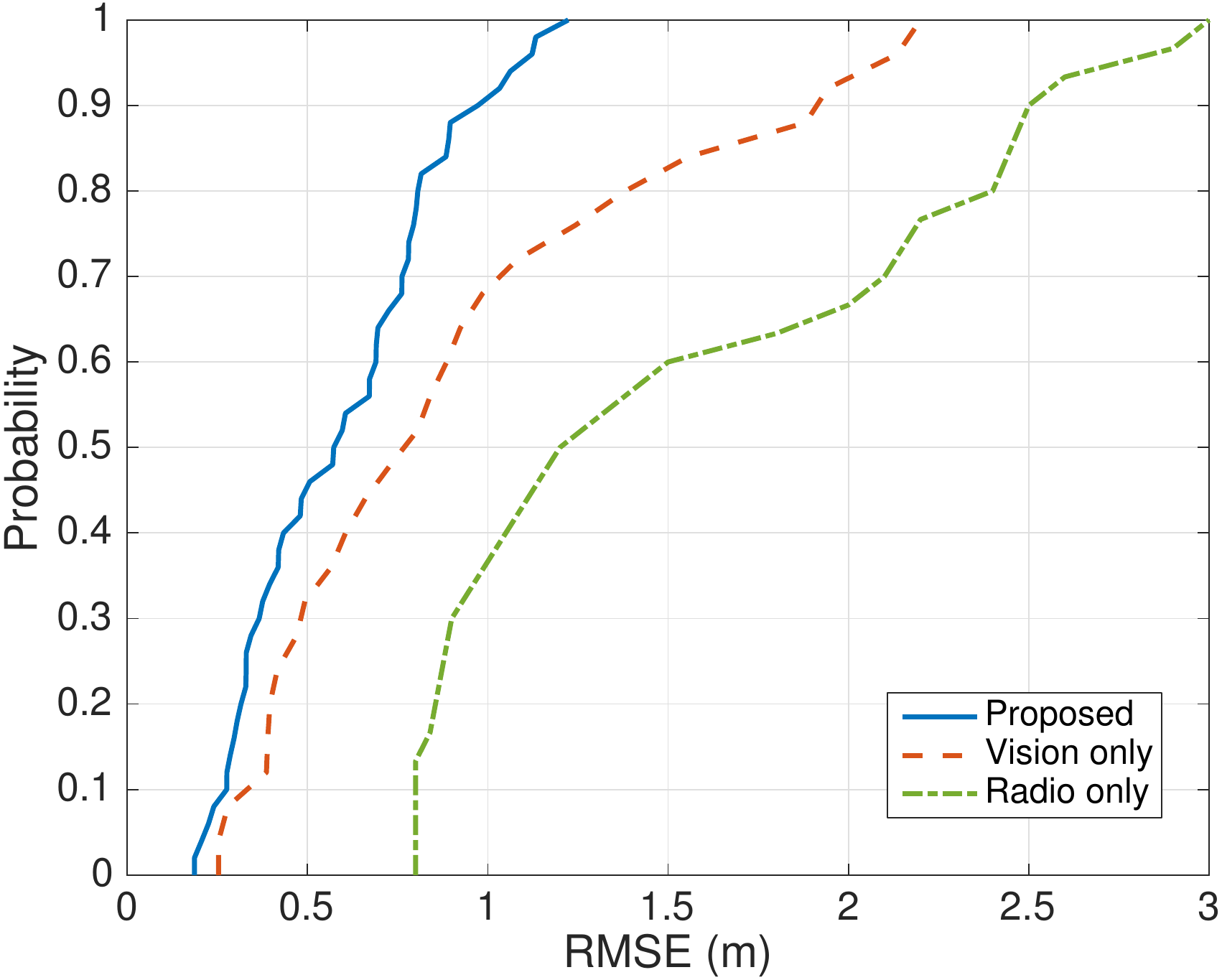}
		\caption{}%
		\label{fig:visionVsProp}%
	\end{subfigure}\hfill%
	\caption{ (a) Cumulative distribution function of RMSE for different learning settings. (b) Accuracy comparison of the proposed approach and the original RBMCDA (vision only) algorithm.}
	\label{}
\end{figure}



\subsection{Results}

\textbf{Accuracy and learning:}
The first set of experiments evaluates the tracking accuracy of our system (i.e. how well we can identify and track people with smartphone devices among all people in the FOV). Moreover, we examine what is the effect of cross-modal training on the performance of our system. Our performance metric in this experiment is the root mean square error (RMSE) between the ground-truth and the estimated trajectory. In all the experiments shown here we have used 100 particles. In addition, instead of using line 13 of Alg. (\ref{alg:rbmcda}) to estimate the filtering distribution, in each step the location of each target is estimated using the particle with the highest weight. For this test we used 30 minutes worth of data running our filter on time-windows of one minute (i.e. 1800 frames). Figure~(\ref{fig:cdf}) shows the error CDF over this period over all targets for different settings. More specifically, our approach achieves a 90 percentile error of 2.0m when the system is untrained, which improves to 1.8m when the foreground detector is trained. The error decreases further as the parameters of the radio propagation model are learned, achieving a 90 percentile error of 1 meter. Finally, once the optimum step-length of each person is learned the accuracy increases further to approximately 0.8 meters. As we can see the error decreases significantly once both the foreground detector and the radio model are learned.
This is expected since our system requires both camera and radio measurements in order to determine the correct measurement to track association and update the target states. In the case of excessive missing camera detections, the trajectory of a target is estimated only by inertial measurements which is the main cause of the low accuracy. On the other hand, if the radio model was not trained, camera detections would not be able to be linked with radio measurements, which would also cause identification and tracking errors. Once the foreground detector and the radio model are trained Fig. (\ref{fig:cdf}) does not show any significant improvement after learning the step-length model. This is reasonable since, in this case most of the time the targets are updated with camera observations which are used to correct the predicted by the IMU states. However, from our experiments we have observed that once the camera becomes unavailable, the difference in accuracy between a trained and a universal step-length model is significant.

\begin{figure}
	\centering
	\includegraphics[width=\columnwidth]{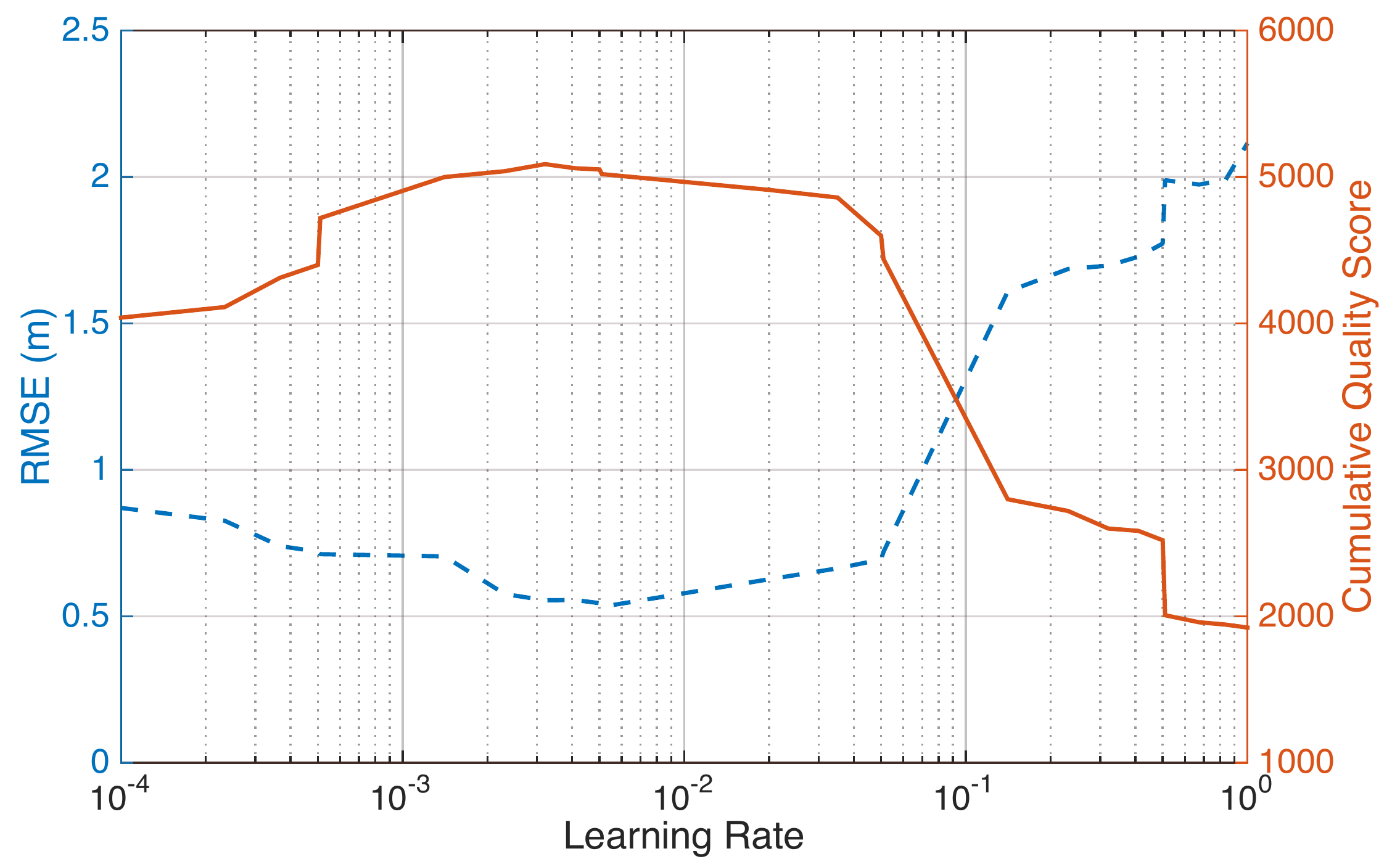}
	\caption{The figure shows the cumulative quality score (CQS) over a period of time as a function of the foreground detector learning rate ($\alpha$). The optimum learning rate according to RMSE maximizes CQS, thus this metric can be used to train the foreground detector.}
	\label{fig:learningRate}
\end{figure}

Figure (\ref{fig:learningRate}) shows how our approach can find the optimum learning rate $(\alpha^*)$ of the foreground detector by solving the optimization problem discussed in Section 6.2. In the example above we used 5 minutes of data, running the foreground detector for different values of $(\alpha)$ and calculating the cumulative quality score (CQS) for that period.  Our intuition is that the optimum learning rate will reduce the number of missing detections, thus increasing the number of high quality tracks as well as their quality score.  This is shown in Fig. (\ref{fig:learningRate}) where the optimum learning rates achieve a high CQS, also evident by the low RMSE.

\begin{figure}
	\centering
	\includegraphics[width=\columnwidth]{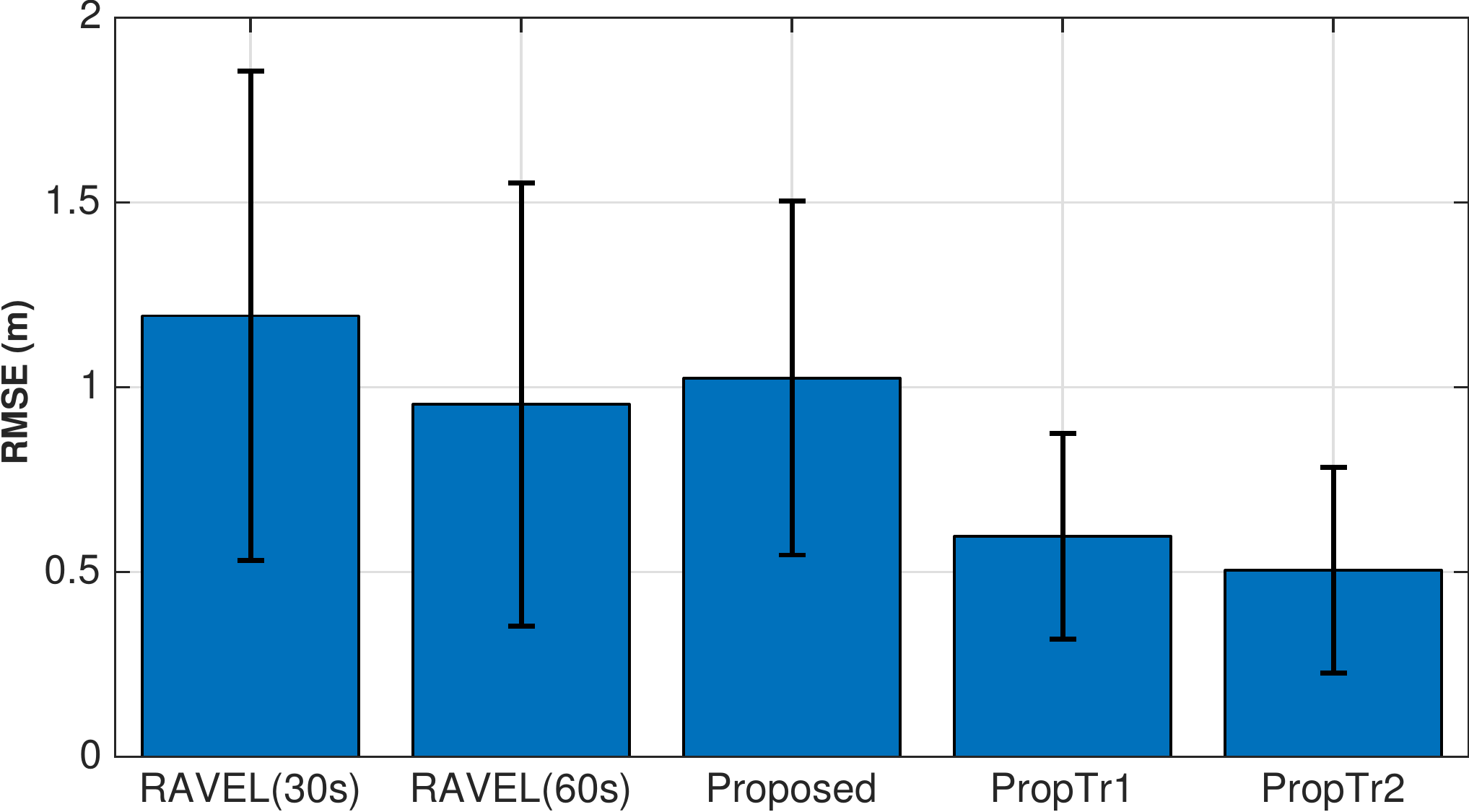}
	\caption{Tracking accuracy between the proposed approach and RAVEL. \emph{Proposed} denotes our approach where the foreground detector and step-length model are not trained. \emph{PropTr1} is our approach after the foreground detector has been trained and further in \emph{PropTr2} the step-length model is also trained. \emph{RAVEL(30s)} and \emph{RAVEL(60s)} is the competing technique evaluated at window sizes of 30 and 60 seconds respectively.}
	\label{fig:test}
\end{figure}


\noindent\textbf{Comparison with other techniques:}
In our second test we compare the proposed approach with the original RBMCDA algorithm (referred to as vision-only tracker in this section)  which uses only visual observations for tracking. In this test we used the same experimental setup as described in the previous paragraph. Both techniques use the same foreground detector settings and in addition the proposed method uses a learned radio model. Figure~(\ref{fig:visionVsProp}) shows the error CDF for the two methods. As we can observe the proposed technique achieves a 90 percentile error of 1 meter as opposed to vision-only tracking which has a 90 percentile error of 1.8 meters. The main source of error for the vision-only tracking is due to data association ambiguities which the proposed technique reduces significantly with the help of radio and inertial measurements. Moreover, the proposed technique supports target identification which is not possible when pure visual tracking techniques are used. In addition, Figure~(\ref{fig:visionVsProp}) shows how the proposed technique stacks up against WiFi fingerprinting. For comparison we have implemented the continuous space estimator of the Horus \cite{Youssef2005}  fingerprinting system (termed as Radio only) by taking into account the 12 WiFi access points in the construction site environment. Figure (\ref{fig:visionVsProp}) gives us a quite good idea of how the WiFi fingerprinting approach performs compared to the proposed system. The 90 percentile error of the radio only technique is approximately 2.5 meters compared to the 1 meter accuracy that the proposed technique achieves.

\begin{figure}
	\centering
	\begin{subfigure}{0.5\columnwidth}
		\includegraphics[width=\columnwidth]{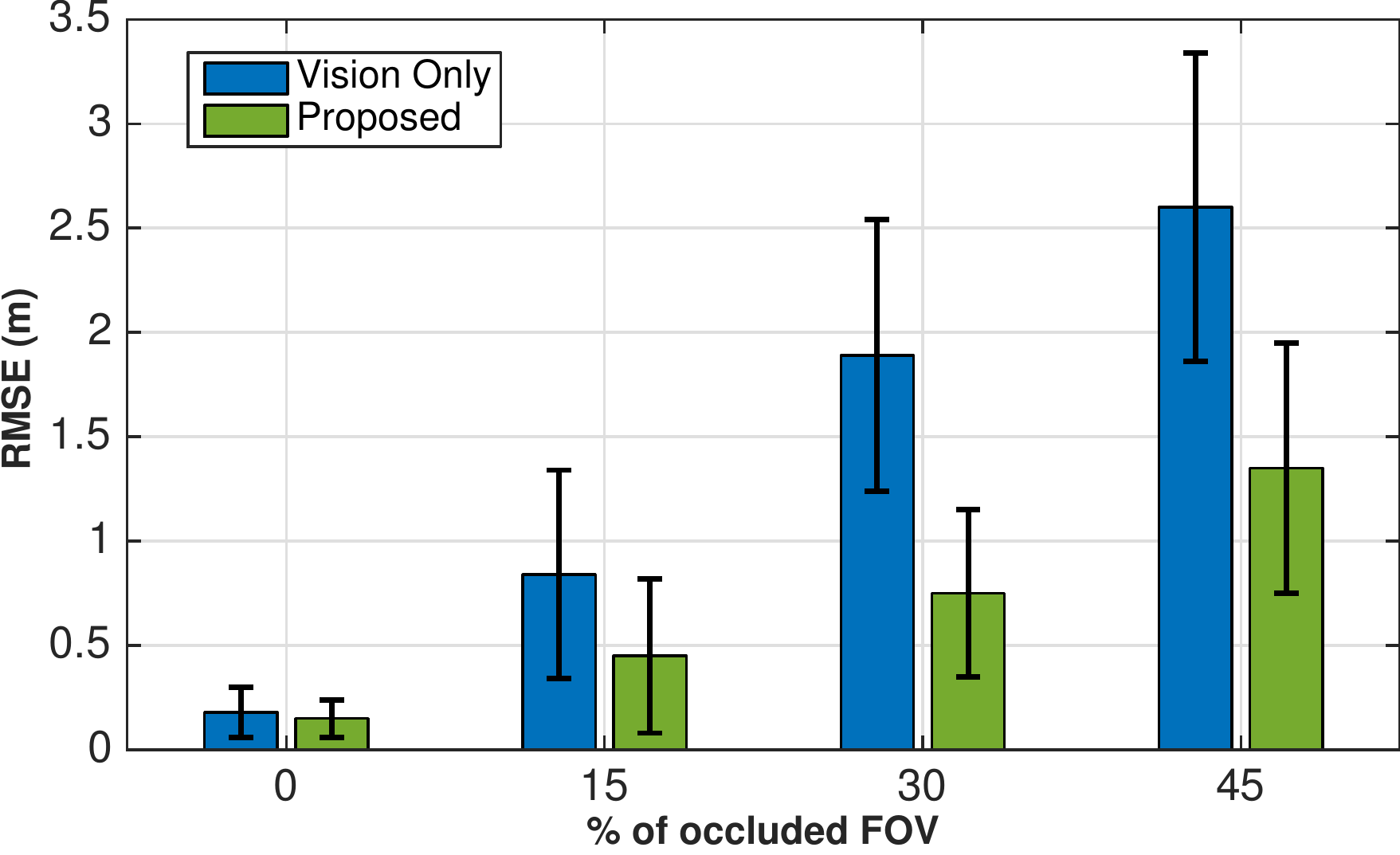}
		\caption{}%
		\label{fig:missDetect}%
	\end{subfigure}\hfill%
	\begin{subfigure}{0.45\columnwidth}
		\includegraphics[width=\columnwidth]{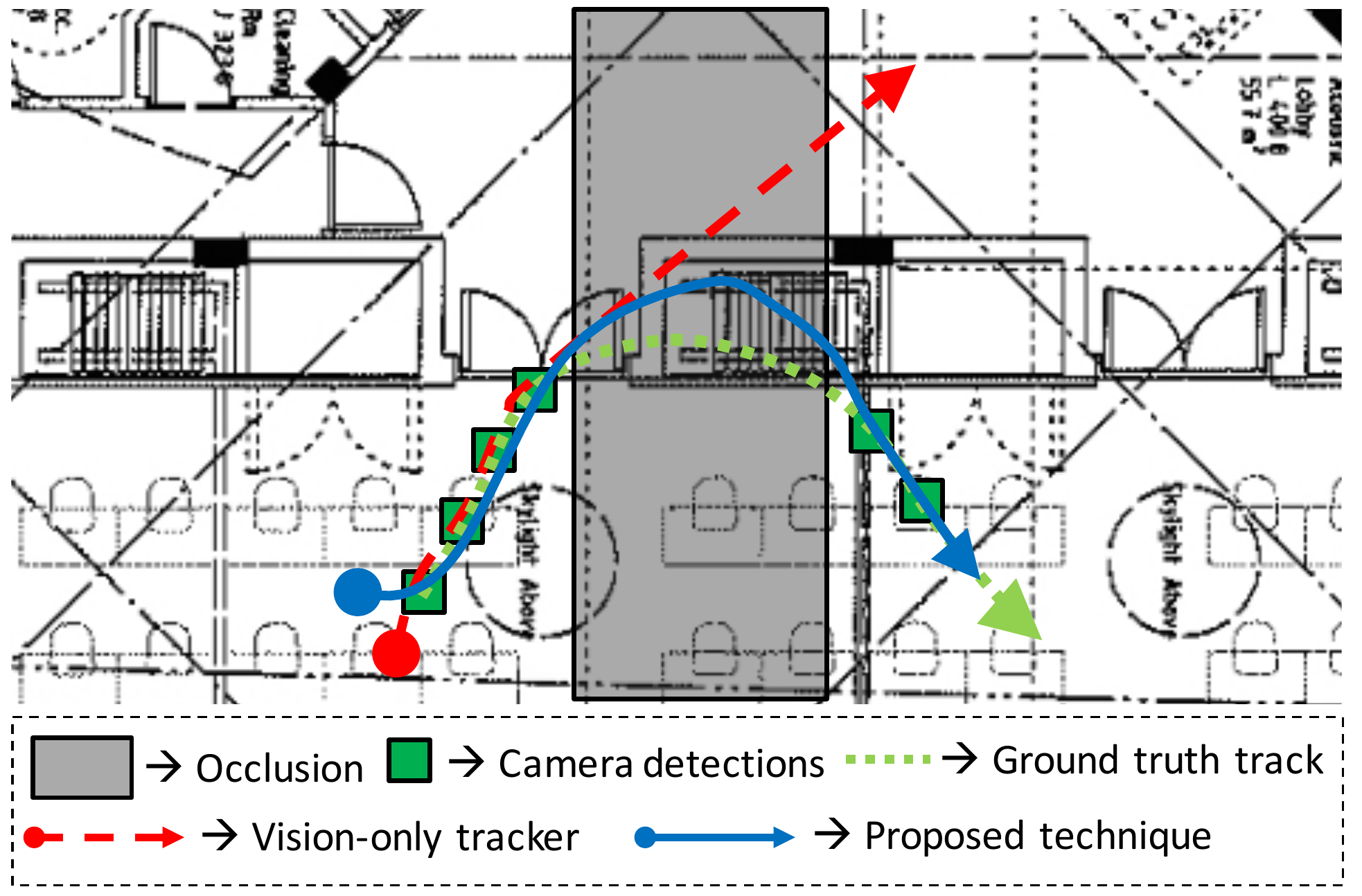}
		\caption{}%
			\label{fig:occlusions}%
	\end{subfigure}\hfill%
	\caption{ (a) The figure shows the RMSE between the proposed technique and the vision-only tracking for different amounts of occlusion. The use of inertial measurements by the proposed technique improves tracking significantly in noisy scenarios. (b) Illustrative example showing the difference between vision-only tracking (red line) and the proposed approach (blue line) in the presence of occlusions (gray area). In cases of prolonged missing camera detections (green squares) the constant velocity model of the vision-only tracker is not sufficient enough to maintain tracking. On the other hand the proposed technique with the aid of inertial measurements is capable of closely following the target despite the presence of long-term occlusions. }
	\label{}
\end{figure}



\noindent The next step is to compare our technique with the recently proposed RAVEL system \cite{Papaioannou2014} which is also a multiple hypothesis tracking and identification system. RAVEL which is discussed in more detail in Section 8 exploits the smoothness of motion and radio signal strength data in order to track and identify targets. Unlike our technique, RAVEL is more of a  reconstruction technique (i.e. performs off-line tracking) as it requires to observe all measurements over a time window ($W$) in order to provide the trajectories of each target. We have tested RAVEL using time windows of sizes 30 and 60 seconds over a period of 10 minutes and we have compared it with the proposed online system. Both systems are capable of learning the radio model parameters, thus we performed these tests using the learned radio model for both systems. 
In Fig. (\ref{fig:test}) \emph{RAVEL(30s)} and \emph{RAVEL(60s)} shows the accuracy of RAVEL for window sizes of 30 and 60 seconds respectively. \emph{Proposed} denotes the proposed system with learned radio model, \emph{PropTr1} is the proposed system optimized one level further i.e. foreground detector training and \emph{PropTr2} denotes the proposed approach when the step-length model is also learned.
Fig. (\ref{fig:test}) shows that the average error of RAVEL decreases from 1.2m to 0.9m as we increase the window size. Our approach with a trained radio model is slightly worse than RAVEL(60). However, once our system trains the foreground detector, the average error decreases significantly and continues to decrease as the step-length model is also learned. Unlike our system, RAVEL estimates the trajectory of a target using only visual data thus it becomes easily susceptible to errors due to missing camera detections. Our system without training achieves a similar performance but in real-time. 

\begin{figure}
	\centering
	\includegraphics[width=\columnwidth]{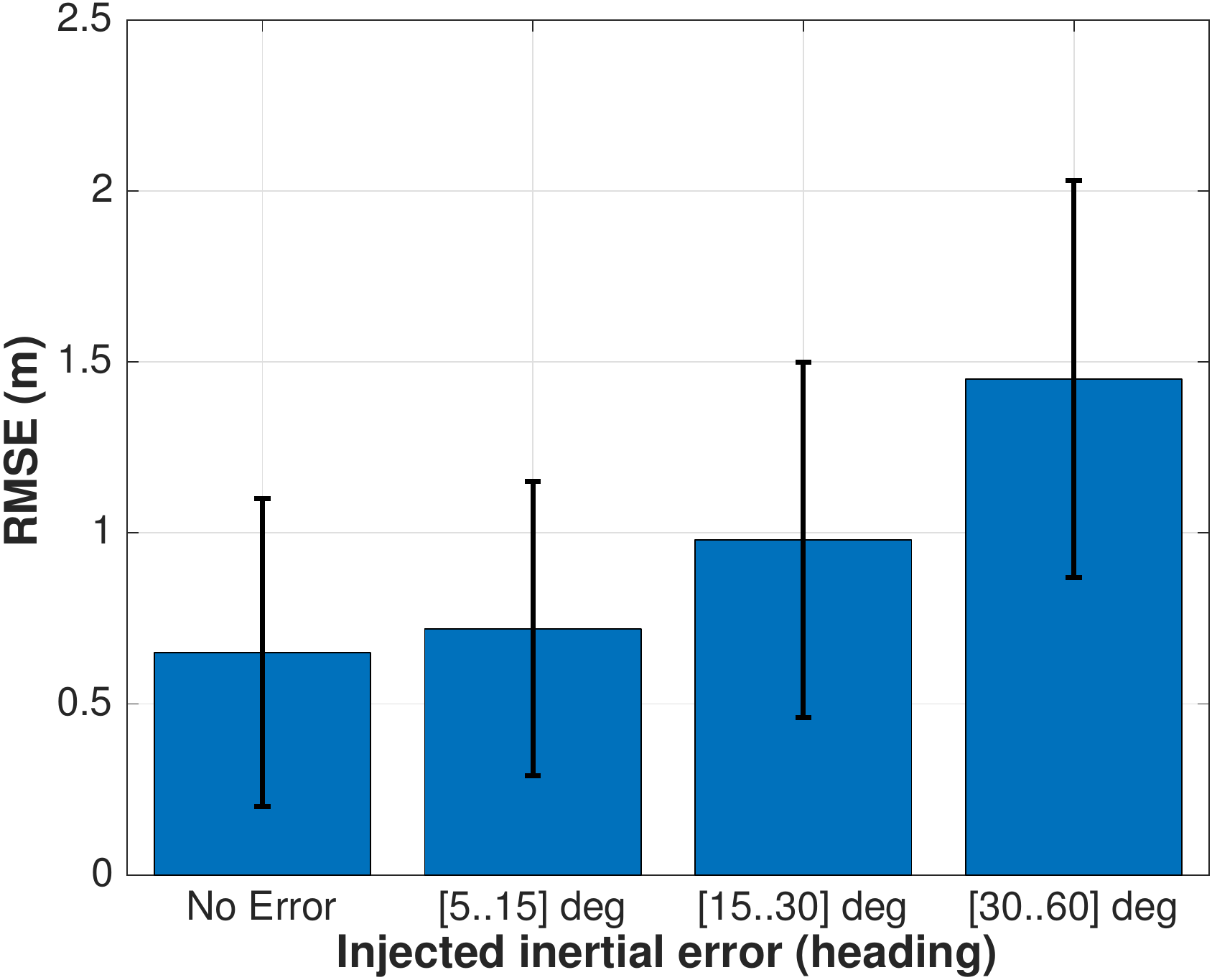}
	\caption{The RMSE of the proposed technique under different amounts of injected heading error.}
	\label{fig:Ierror}
\end{figure}


\noindent \textbf{Robustness:}
This set of experiments aims to demonstrate the robustness of the proposed technique. First we wanted to see how our technique performs on difficult trajectories (i.e. various amounts of occlusions and missing detections). In order to simulate occlusions we remove a specific area of the field of view (FOV) by disabling the camera detections inside that area. More specifically, we generated occlusions at random locations that occupy a rectangular area of specific size inside the FOV. Then we evaluated the accuracy of the proposed approach compared to the vision-only tracker on 50 trajectories of variable length generated from our ground truth data. Fig. (\ref{fig:missDetect}) shows  the RMSE over all trajectories between the proposed system and the vision-only tracker for different configurations of occlusions (i.e. shown as the percentage of occluded FOV). For each configuration we run the test 10 times; each time the occlusion was positioned to a different location. 
The two methods achieve a comparable performance when there are no occlusions. However, the proposed approach significantly outperforms the vision-only tracking in scenarios with long-term occlusions and large amounts of missing detections. In the presence of long-term occlusions the constant velocity/acceleration motion model utilized by most visual tracking techniques fails and cannot be used to reliably model the inherently complex human motion. On the other hand Fig. (\ref{fig:missDetect}) shows that the use of inertial measurements by the proposed technique provides a more accurate model of human motion. An illustrative example is shown in Fig. (\ref{fig:occlusions}).

Additionally in order to study how our approach can cope with variable noise from the inertial sensors we followed a similar procedure as in the previous paragraph and we generated 50 trajectories from our ground truth data. At each time-step and for each trajectory we inject a random bias error to the heading estimator. More specifically we sample a heading error uniformly from a specific range of the form $[a..b] ~degrees$  and we add it to the output of the heading estimator. By doing this we can get an idea of how our approach performs in environments with disturbed magnetic fields. Fig. (\ref{fig:Ierror}) illustrates the results of this experiment for different amounts of injected noise. As we can see the proposed technique can cope with moderate amounts of inertial noise; achieving a sub-meter accuracy for bias up to 30 degrees.

\begin{figure}
	\centering
	\includegraphics[width=\columnwidth]{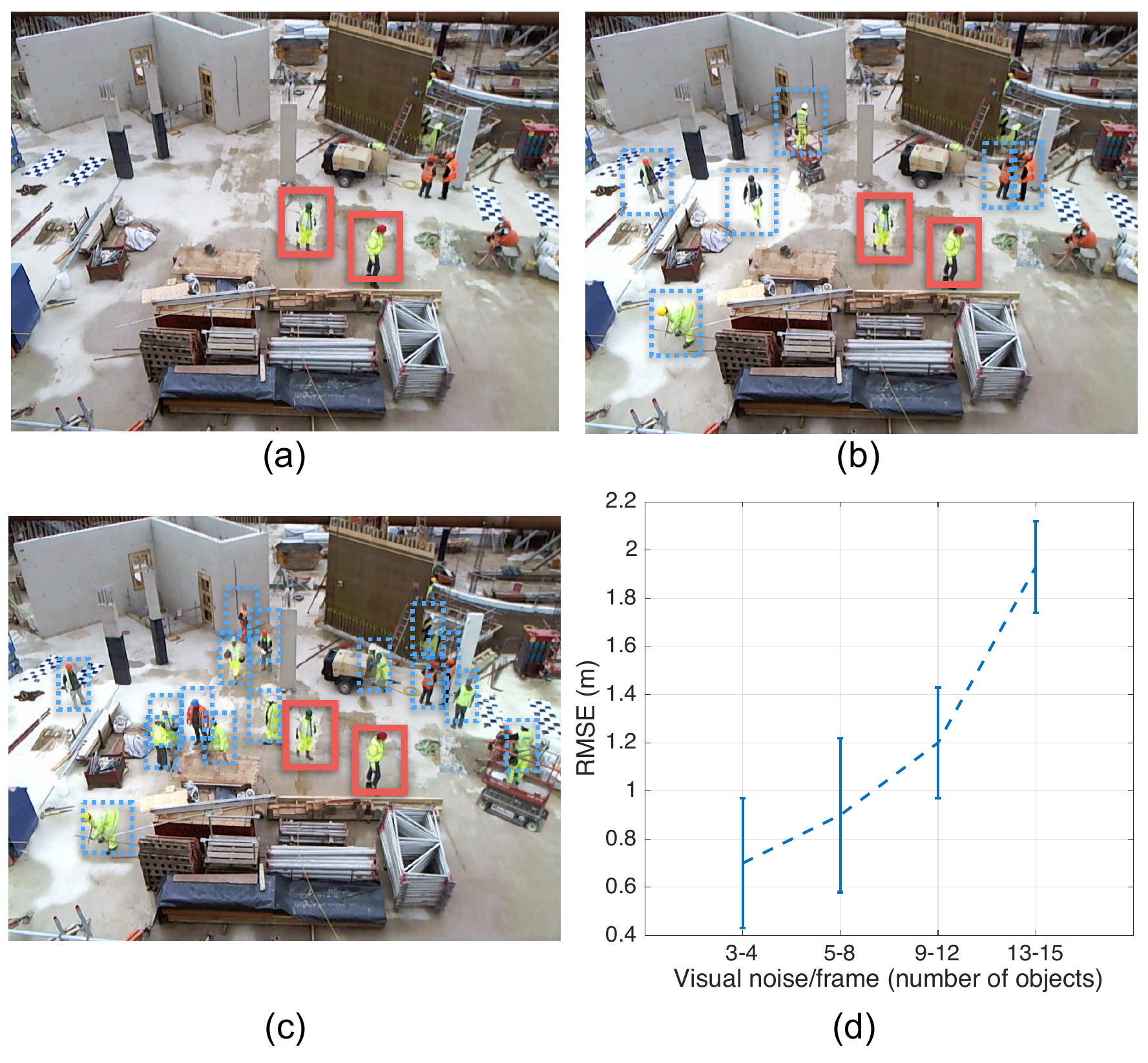}
	\caption{The RMSE of the proposed technique under different amounts of visual noise. (a) Camera snapshot without visual noise where we track the people in red rectangles. (b) Visual noise is injected in the scene (i.e. objects in blue rectangles). (c) Additional visual noise is injected in the scene. (d) The impact of visual noise on the performance of the proposed approach.}
	\label{fig:visualNoise}
\end{figure}

Moreover, we wanted to see how the number of people in the scene affects the performance of our system and in addition what is the impact of visual noise on the tracking accuracy. In order to study this, we used 10 minutes worth of data (i.e. 18000 frames) from our construction site dataset. For each frame in this dataset we have superimposed visual objects from future timestamps in order to increase the visual noise and the number of people in the scene. We have split the dataset in windows of 1 minute each (i.e. 1800 frames) and we recorded the RMS error for different number of visual objects as shown in Fig. (\ref{fig:visualNoise}). It is worth noting that, the number of people that we track includes only the people which carry mobile devices (i.e. 5 people). As we can see from Fig. (\ref{fig:visualNoise}) as we increase the number of visual objects in the scene the accuracy drops. More specifically when we have relatively small number of objects in the scene (i.e. 3-4 per frame) the error is approximately 0.7 meters and increases to approximately 1.9 meters when the number of objects increases to 13-15 per frame. The reason behind this is due to the fact that the WiFi cannot distinguish close-spaced targets and despite the use of IMU data for motion prediction a track can incorrectly be updated with the wrong visual observation (i.e. visual noise).  Possible solutions to this problem, is to consider the evolution of the WiFi signal over multiple frames as opposed to the on-line filtering approach that we have currently implemented. Additionally multiple overlapping cameras can also help but this will increase the system's cost and complexity.
	

\begin{figure}
	\centering
	\begin{subfigure}{0.5\columnwidth}
		\includegraphics[width=\columnwidth]{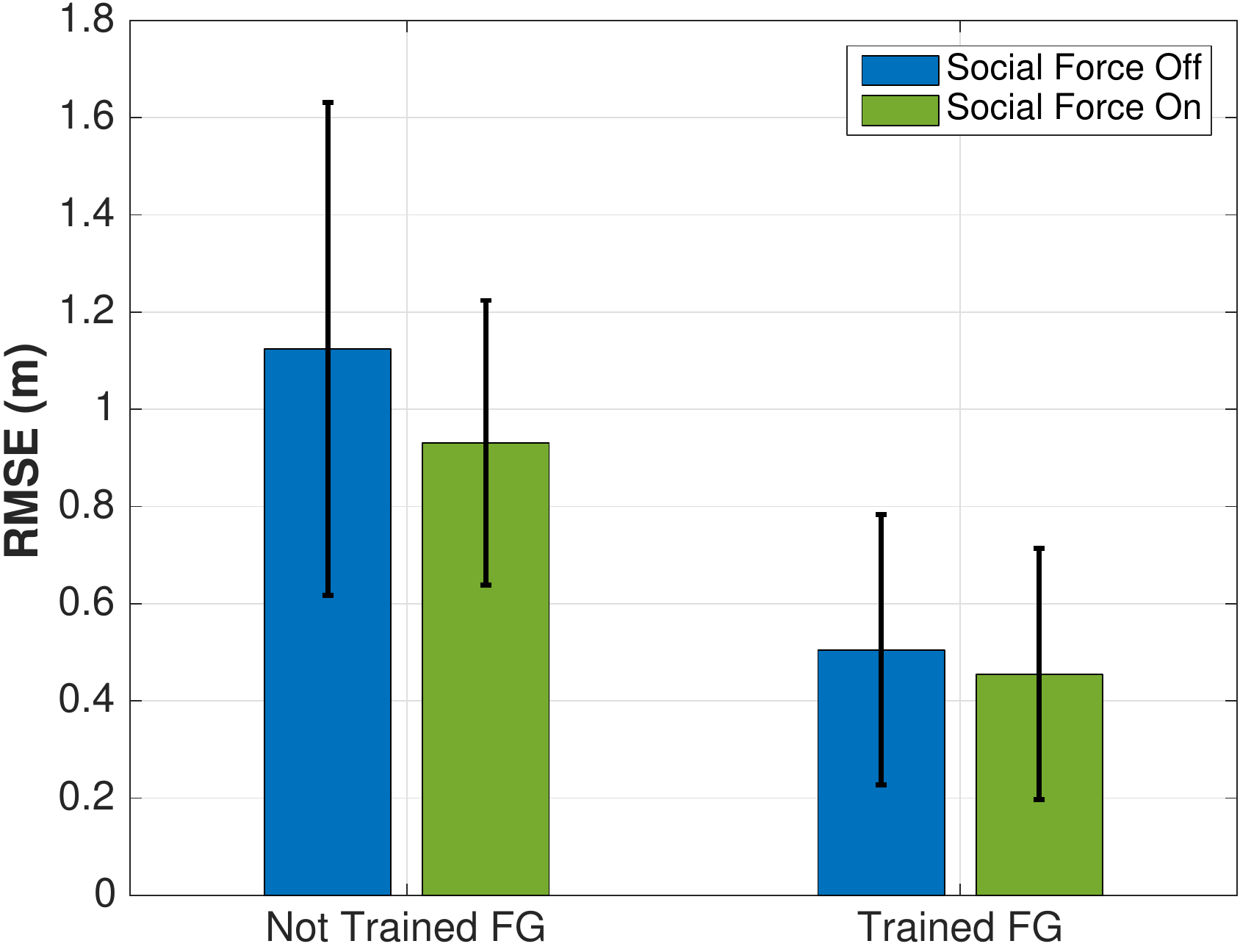}
		\caption{}%
		\label{fig:socialForceTest1}%
	\end{subfigure}\hfill%
	\begin{subfigure}{0.5\columnwidth}
		\includegraphics[width=\columnwidth]{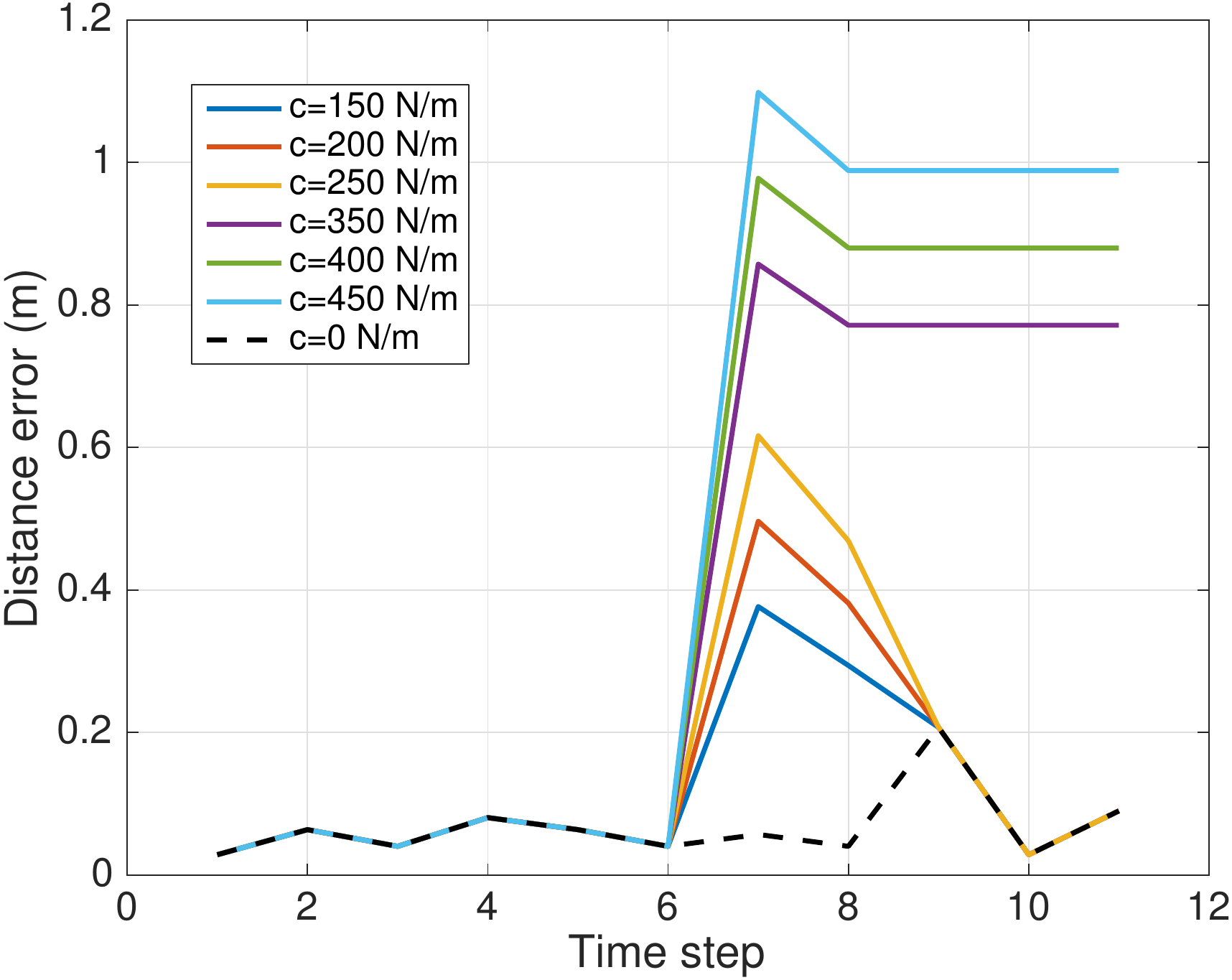}
		\caption{}%
		\label{fig:socialForseTest2}%
	\end{subfigure}\hfill%
	\caption{ (a) Impact of Social Forces on the performance of our system. (b) Tuning the parameters of the social force model. The graph shows the impact of the force magnitude ($c_j $) from Eq. (\ref{eq:physForce}) on the accuracy of the system.}
	\label{}
	\vspace*{-5 mm}
\end{figure}

\noindent \textbf{Impact of social forces:}
The last set of experiments aims to investigate the impact of social forces on the performance of our system. For this experiment we used our improved motion model given by Eqn. (\ref{eq:forceMotion}) that takes into account the influences of people and the environment on human motion. Two tests were conducted; first we have investigated the scenario where the visual detector (i.e. foreground detector in our case) does not perform optimally and so the missing camera detection rate is high. The second test deals with a trained visual detector. In the first case, our system will rely mostly on inertial measurements. Our intuition is that the addition of social forces will improve the motion prediction; thus increasing the overall tracking accuracy. Social forces essentially help us avoiding predictions through obstacles (i.e. walls) and also help us model the interactions between people. Figure (\ref{fig:socialForceTest1}) shows the results of this test which is based on 15 minutes worth of data, where we compare the impact of social forces on two different settings of the foreground detector (i.e. not trained and trained). In this test we assume that people have a mass of 70Kg and a radius of 0.2m. The rest of SFM parameters are as follows $a_j=50$N, $b_j=0.5$m, $\lambda=0.5$ and $c_j=250$N/m. As we can see the social forces improve the overall accuracy by approximately 20\% on a non-trained foreground detector and the improvement on a trained foreground detector is roughly 10\%. The reason behind these improvements is due to the more accurate motion prediction which allows a person to move more accurately in the environment even without camera detections. As the environment is populated with more constraints (i.e. walls, corridors) the gain of using the SFM is increasing. A second reason for these improvements is due to the fact that now the predicted locations are more aligned with the actual observations which improves the final position estimates and in addition reduces the data association errors.

Finally, we should note that selecting correctly the parameters of the social force model is very important if we would like the SFM to be beneficial and improve the tracking accuracy. Figure. (\ref{fig:socialForseTest2}) shows the impact of 
the force magnitude ($c_j $) from Eq. (\ref{eq:physForce}) on the accuracy of the system in the case of an erroneous obstacle (i.e. we have incorrectly estimated the presence of an obstacle, when in fact it does not exist).  More specifically, in this example we assume that an erroneous obstacle is blocking the trajectory of a person. This obstacle exerts physical forces to this person in order to restrict his motion. During time steps 1 to 5 the obstacle is far away and the social force has no effect on the human motion. However, when the person is close enough  (e.g. time step 7), the social force exerted onto him is opposite to the direction of his motion (this is to prevent a person to go through the obstacle). As we increase the force magnitude ($c_j$) the error from the ground truth (i.e. $c_j=0$ N/m) increases since this increasing force is pushing the person further away. Now, for some values of $c_j $ (e.g. 150-250 N/m) the acting force has the right magnitude and allows a person to go through the obstacle in cases where we have measurements on and beyond the obstacle area. However, when the force is too large (i.e. 450 N/m) , a person cannot go through the obstacle and in the scenario of an erroneous obstacle the correct path (i.e. $c_j=0$ N/m) cannot be recovered as shown in the graph. We have found experimentally that the SFM works best if it is tuned so that it would point towards the right direction but without causing significant repulsion. This strategy allows us to have improved location predictions that align better with the actual observations but also allows targets to go through obstacles/occlusions in cases of incorrect obstacle inference. 


\section{Related Work}\label{sec:background}

A variety of positioning systems have been proposed by the research community over the past ten years. Recent surveys outlining the different techniques and their accuracies can be found in  \cite{ipsnComp,Mautz:IPS-Book:2012}. In this section we will give a brief overview on the most recent positioning systems that make use of radio-, inertial- and visual- sensing ( i.e. using a stationary camera) to track multiple people. The positioning systems to be described here can be divided into two categories: a) systems that combine visual and radio measurements and b) those that combine visual and inertial measurements.

\noindent \textbf{Vision+Radio positioning systems:} The Radio And Vision Enhanced Localization (RAVEL) system \cite{Papaioannou2014} fuses anonymous visual detections captured by a stationary camera with WiFi readings to track multiple people moving inside an area with CCTV coverage. The WiFi measurements of each person are used to add context to the trajectories obtained by the camera in order to resolve visual ambiguities (e.g. split/merge paths) and increase the accuracy of visual tracking. RAVEL operates in two phases namely tracklet generation and WiFi-aided tracklet merging. In the first phase visual detections collected over a period of time are used to form unambiguous small trajectories (i.e. tracklets). In the second phase, RAVEL uses the aforementioned tracklets to create tracklet trees for each person (i.e. probable trajectory hypotheses). Then, the WiFi measurements of each person are used to search through the tracklet tree in order to find their most likely trajectory. The most likely trajectory is the one that agrees the most with the WiFi measurements. Unlike our technique, RAVEL performs off-line tracking, i.e. the trajectory of each person is reconstructed after all camera detections and WiFi measurements for a period of time have been observed. In addition, RAVEL does not make use of inertial measurements and thus it is more susceptible to positioning errors due to missing detections (i.e. static people that become part of the background).

In a similar setting the EV-Loc system \cite{Zhang2012} estimates the position of multiple people using both WiFi and camera measurements. More specifically, EV-Loc estimates the distance of each person from a number of access points first using camera measurements and then using WiFi readings. The Hungarian algorithm \cite{bourgeois1971,Kuhn1955} is then used to find the best mapping between camera and WiFi measurements. After this optimization problem is solved, the camera and WiFi locations of each person are fused to form a weighted average final location. Unlike our work, EV-Loc concentrates on the problem of finding the best matching between camera and WiFi traces (i.e. the matching process is performed after the visual tracking is completed ) and does not provide a general tracking framework that incorporates multiple sensor modalities. The more recent RGB-W system \cite{Alahi2015} also uses wireless signals emitted by people's mobile phones in combination with cameras to track and identify people. The authors show how the wireless signals can be used as a rough proxy for depth information which allows them to achieve better localization accuracy.

Mandeljc et al. presented in \cite{Mandeljc2011,Mandeljc2012} a fusion scheme that extends the probabilistic occupancy map (POM) \cite{Fleuret2008} with radio measurements. 
In \cite{Mandeljc2011} the POM is extended so that the cell occupancy probabilities are estimated using ultra-wideband (UWB) radio sensors in addition to the cameras. 
This additional radio information increases the accuracy and robustness of the algorithm. Later in \cite{Mandeljc2012}, the POM is extended further so that the anonymous camera detections are augmented with identity information from radio tags. The augmentation of anonymous detections with identity information is done on a frame-by-frame basis where at each time instant the optimal assignment between radio and camera locations is obtained using the Hungarian algorithm. The fusion scheme of \cite{Mandeljc2011,Mandeljc2012} was evaluated using only UWB radios which exhibit sub-meter accuracy and there is no indication of how this method will perform with radios of lower accuracy (i.e. WiFi). 
Finally, in \cite{Goller2014} Goller et al. presents a hybrid RFID and computer vision system for localization and tracking of RFID tags. The authors show increased accuracy by combining the two complimentary sensor modalities in a probabilistic manner.

\noindent \textbf{Vision+Inertial positioning systems:} Instead of using radio measurements for identification the methods in this category use inertial measurements. For instance, the system in \cite{Teixeira2009} fuses motion traces
obtained from one stationary camera mounted on the ceiling and facing down with motion information from wearable accelerometer nodes to uniquely identify multiple people in the FOV using their accelerometer node IDs. Background subtraction is used to detect people from the video footage and then their floor-plane acceleration is extracted by double differentiation. The camera acceleration traces are then compared against the overall body acceleration obtained from the accelerometer nodes using the Pearson's correlation coefficient. The acceleration correlation scores among all possible combinations of camera-accelerometer pairs are then used to form an assignment matrix. Finally, the assignment problem is solved using the Hungarian algorithm. The initial algorithm of \cite{Teixeira2009} is extended in \cite{Jung2010} to allow for better path disambiguation based on people's acceleration patterns by keeping track of multiple trajectory hypotheses.

\section{Future Work}
In this paper we have presented a novel tracking system that uses three different sensor modalities (i.e. visual, radio and inertial) to accurately track and identify multiple people in a construction site setting. In addition we have developed learning techniques that make the proposed system able to adapt to the highly dynamic environment of the construction site. So far in our system we used a single stationary camera in order to monitor and track the people in the scene. Our next step is to extend our system to use multiple cameras in order to provide location services to larger areas. 
	
Since the proposed technique is able not only to track but also to identify the people a simple approach would be to replicate and deploy the existing system in different areas (i.e. each deployment would use a single camera). However, we believe that better performance can be achieved by considering the collaboration between different cameras. The next step is to extend the proposed technique to a multi-camera multi-target tracking system by taking into account transition probabilities between multiple non-overlapping cameras. So far we have covered the case of 2D tracking in large unconstrained/open areas. As a future step we will also consider extending the current system to cover tracking in 3D. 

Furthermore, building a stable network that can support such a system is also a challenging task. We need to think about the required network bandwidth, efficient communication between the different sub-systems and synchronization. However, all the above are going to be investigated in our future work.
\section{Conclusion}
In this paper we proposed a multi-modal positioning system for highly dynamic environments. We showed that it is possible to adapt Rao-Blackwellised particle filters - traditionally used to discern tracks using anonymous measurements - in order to both identify and track people being monitored by CCTV and holding mobile devices. We further showed that there is significant scope for automatically training the various sensor modalities, and this proved particularly useful in rapidly changing environments. Additionally, we showed that the use of social forces in dynamic industrial environments is highly beneficial and improves the tracking accuracy.
Our experiments showed that even without training, our online approach achieves similar positioning accuracy to the existing off line RAVEL approach; with training the positioning error is decreased by a further 50\%. We also showed that the proposed technique is robust in scenarios with visual and inertial noise. Lastly, with the integration of social forces we improved the accuracy by 10-20\%.

\section{Acknowledgments}
We would like to thank Laing O'Rourke  for allowing us to conduct our experiments in their construction site and also for funding this research. 

\bibliographystyle{IEEEtran}
\bibliography{IEEEabrv,bib} 
\newpage
\begin{IEEEbiography}[{\includegraphics[width=1in,height=1.25in,clip,keepaspectratio]{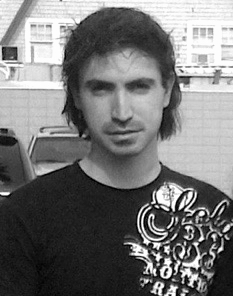}}]%
{Savvas Papaioannou} is currently a PhD student in the Computer Science Department at University of Oxford. He obtained a B.S. degree in Electronic and Computer Engineering from Technical University of Crete and a M.S. degree in Electrical Engineering from Yale University. His research interests focus on sensor networks, localization and multiple target tracking.
\end{IEEEbiography}
\balance
\begin{IEEEbiography}[{\includegraphics[width=1in,height=1.25in,clip,keepaspectratio]{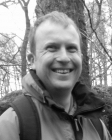}}]%
{Dr. Andrew Markham} received his BSc.Eng (Hons) degree in 2004 and his PhD in 2008 from the University of Cape Town, South Africa. He is currently an Associate Professor in Software Engineering at the Department of Computer Science, University of Oxford. His primary research interests are in low power sensing and communication.
\end{IEEEbiography}

\begin{IEEEbiography}[{\includegraphics[width=1in,height=1.25in,clip,keepaspectratio]{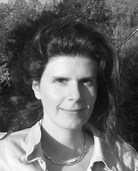}}]%
{Dr. Niki Trigoni} is a Professor at the Department of Computer Science, University of Oxford. She obtained her PhD at the University of Cambridge (2001), became a postdoctoral researcher at Cornell University (2002-2004), and a Lecturer at Birkbeck College (2004-2007). Since she moved to Oxford in 2007, she established the Sensor Networks Group, and has conducted research in communication, localization and in-network processing algorithms for sensor networks. Her recent and ongoing projects span a wide variety of sensor networks applications, including indoor/underground localization, wildlife sensing, road traffic monitoring, autonomous (aerial and ground) vehicles, and sensor networks for industrial processes. 
\end{IEEEbiography}

\end{document}